\documentclass{article} 
\usepackage{iclr2027_conference,times}

\usepackage{amsmath,amsfonts,bm}

\def\eqref#1{equation~\ref{#1}}

\def\1{\bm{1}}

\DeclareMathAlphabet{\mathsfit}{\encodingdefault}{\sfdefault}{m}{sl}
\SetMathAlphabet{\mathsfit}{bold}{\encodingdefault}{\sfdefault}{bx}{n}

\usepackage{hyperref}
\usepackage{url}
\usepackage{graphicx}
\usepackage{algorithm}
\usepackage{algpseudocode}
\usepackage{xcolor}
\usepackage{amssymb}
\usepackage{booktabs}
\usepackage{multirow}
\usepackage{wrapfig}
\usepackage{enumitem}

\usepackage{capt-of}
\usepackage{pifont}
\usepackage{colortbl}

\usepackage{needspace}

\algtext*{EndIf}
\algtext*{EndWhile}

\definecolor{RIRWhen}{HTML}{8B6555}
\definecolor{RIRWhere}{HTML}{8B6555}
\definecolor{RIRWhat}{HTML}{8B6555}

\definecolor{rirhi}{HTML}{EDF1F8}

\newcommand{\WhenTag}[1]{%
    \Comment{\textcolor{RIRWhen}{%
    \scriptsize \textsc{When}\,$\cdot$\,#1}}%
}

\newcommand{\WhereTag}[1]{%
    \Comment{\textcolor{RIRWhere}{%
    \scriptsize \textsc{Where}\,$\cdot$\,#1}}%
}

\newcommand{\WhatTag}[1]{%
    \Comment{\textcolor{RIRWhat}{%
    \scriptsize \textsc{What}\,$\cdot$\,#1}}%
}

\usepackage[listings,skins]{tcolorbox}
\usepackage{listings}
\tcbuselibrary{breakable}

\lstdefinestyle{rirjson}{
  basicstyle=\ttfamily\scriptsize,
  columns=fullflexible,
  keepspaces=true,
  showstringspaces=false,
  breaklines=true,
  aboveskip=0pt, belowskip=0pt,
}

\definecolor{CaseNeutral}{HTML}{3F4550}     
\definecolor{CaseNeutralBG}{HTML}{F4F3EF}
\definecolor{CaseFail}{HTML}{B76565}        
\definecolor{CaseFailBG}{HTML}{F7EBEB}
\definecolor{CaseRollback}{HTML}{C28B55}    
\definecolor{CaseRollbackBG}{HTML}{FBF1E4}
\definecolor{CaseMemory}{HTML}{6E8C6A}      
\definecolor{CaseMemoryBG}{HTML}{EEF3EC}
\definecolor{CaseRecover}{HTML}{2E5596}     
\definecolor{CaseRecoverBG}{HTML}{EAEFF7}
\definecolor{CaseSuccess}{HTML}{2F7A4F}     
\definecolor{CaseSuccessBG}{HTML}{E7F3EA}
\definecolor{CaseBad}{HTML}{8E6C8A}         
\definecolor{CaseBadBG}{HTML}{F3EDF3}
\definecolor{CaseCrit}{HTML}{9C3B3B}        
\definecolor{CaseCritBG}{HTML}{F8EBEB}
 
\newtcolorbox{casestudy}[1]{
  enhanced, breakable,
  colback=white, colframe=black!45,
  boxrule=0.45pt, arc=1.5pt,
  left=6pt, right=6pt, top=5pt, bottom=5pt,
  fonttitle=\bfseries\footnotesize, coltitle=black, fontupper=\small,
  attach boxed title to top left={yshift=-1pt, xshift=2pt},
  boxed title style={boxrule=0pt, colframe=black!7, colback=black!7, arc=1pt},
  title={#1},
}
 
\newtcolorbox{casephase}[4]{
  enhanced,
  colback=#2, colframe=#1,
  boxrule=0pt,
  borderline={0.55pt}{0pt}{#1,dashed},
  arc=1pt,
  left=4pt, right=4pt, top=3pt, bottom=3pt,
  before upper={{\footnotesize\textcolor{#1}{\textbf{#3}}\hfill{\scriptsize\ttfamily #4}}\par\vspace{3pt}},
}
 
\newcommand{\casebadge}[2]{%
  \tcbox[on line, boxsep=1pt, left=3pt, right=3pt, top=1pt, bottom=1pt,
  colframe=#1, colback=#1!10, arc=1pt, boxrule=0.6pt,
  fontupper=\bfseries\scriptsize, coltext=#1]{#2}%
}
 
\title{Rollback the World, Keep the Reflection: Rollback-Induced Reflection for Long-Horizon LLM Agents}

\author{
Yi Yu\textsuperscript{1},
Liuyi Yao\textsuperscript{2,\textdagger},
Yaliang Li\textsuperscript{2},
Enshu Wang\textsuperscript{1},
and Libing Wu\textsuperscript{1,\textdagger}
\\[3pt]
{\normalfont
\textsuperscript{1}Wuhan University
\qquad
\textsuperscript{2}Alibaba Group
}
\\[3pt]
{\normalfont\footnotesize
\texttt{\{yui1212,wanges17,wu\}@whu.edu.cn}
\quad
\texttt{\{yly287738,yaliang.li\}@alibaba-inc.com}
}
}

\iclrfinalcopy 
\begin{document}

\maketitle
\fancyhead[L]{}

\begingroup
\renewcommand{\thefootnote}{\fnsymbol{footnote}}
\footnotetext[2]{Corresponding authors.}
\endgroup

\begin{abstract}
Large language model (LLM) agents increasingly tackle long-horizon tasks through multi-step environment interaction, yet a single erroneous action can alter subsequent states and observations, causing errors to compound over time. Existing methods either correct the context without repairing altered environment states or restore earlier states while discarding useful experience, making it difficult to both eliminate failure conditions and avoid repeating past mistakes. We argue that reliable recovery should instead be treated as a rollback-boundary control problem that jointly determines \emph{when} to intervene, \emph{where} to resume, and \emph{what} information should survive recovery. Based on this view, we propose Rollback-Induced Reflection (RIR), a unified recovery framework that restores execution to a selected prior state while carrying forward reusable knowledge distilled from the abandoned trajectory to guide subsequent decisions. We further characterize recovery through a unified operator over rollback depth and retained memory, providing a general view of state restoration and knowledge retention. Experiments on three long-horizon benchmarks show that RIR consistently improves average task performance across multiple LLM backbones, with structured reflection memory preserving useful experience and selective rollback enabling efficient recovery.
\end{abstract}
\section{Introduction}

Large language model (LLM) agents can solve complex long-horizon tasks through reasoning, tool use, and environment interaction~\citep{wei2022chain,yao2022react,park2023generative}. Yet such tasks are highly sensitive to erroneous actions: a single misstep can alter subsequent states and observations, causing later decisions to rely on corrupted context. Errors therefore compound over time and can drive the agent progressively away from a valid solution trajectory~\citep{hao2026speculative}.

Existing remedies intervene at two levels. \textit{Information-level} methods append corrective feedback to guide future decisions~\citep{madaan2023self,shinn2023reflexion, zhao2024expel,kim2025reflact}, but cannot undo environmental changes already caused by an erroneous action, and contaminated observations remaining in context may contradict the corrective
advice itself. \textit{State-level} methods instead restore execution to an earlier checkpoint and
discard the erroneous suffix~\citep{zhou2023language,li2025generator,zhang2026webrollback,hao2026speculative},
but rewinding the state does not decide which claims from that suffix survive it. Retain too little and the agent may repeat the same failure; overgeneralize a local failure and viable alternatives may be incorrectly ruled out.

A reliable rollback mechanism must therefore answer three coupled questions. First,
\textit{when to roll back}: is the current branch still productive exploration, or has
it begun to propagate errors? Intervening too early curtails valid exploration; too
late lets the consequences of an error compound. Second, \textit{where to roll back}:
which checkpoint eliminates the conditions that produced the failure while preserving
the most valid progress? Going too far back sacrifices completed work; not far enough
leaves the underlying obstacle in place. Third, \textit{what to roll back}: which
claims in the discarded suffix are invalidated along with the execution state, and
which should survive as knowledge for the next attempt?

Together, these questions define a rollback boundary: \textit{when} decides whether
to draw it, \textit{where} places it along the trajectory, and \textit{what} controls
which information crosses it. At this boundary, the system restores the environment
and branch-local agent context to the selected checkpoint and removes the subsequent
suffix from the active trajectory. The suffix, however, need not be discarded
wholesale. State claims invalidated by restoration are removed, while reusable
observations and lessons are distilled into reflective knowledge for the next attempt.
Rollback thus becomes both a state transition and an opportunity to reconstruct the
decision context: \textbf{rollback the world, keep the reflection}.

Building on this view, we propose \textbf{Rollback-Induced Reflection (RIR)}, a
recovery-control framework for long-horizon LLM agents. For \textit{when},
\textbf{Hybrid Adaptive Review} combines agent-initiated and adaptive scheduled reviews
to assess whether the current branch should continue or recover. For \textit{where},
\textbf{Coarse-to-Fine Restore Localization} first narrows the search to a causally
relevant interval and then selects a checkpoint that balances failure removal against
progress preservation. For \textit{what}, \textbf{Rollback-Consistent Reflection Memory} separates branch-local state restored with the checkpoint from reusable knowledge that persists across rollback. The memory stores the stable task objective, reusable environment knowledge, past-attempt milestones, and conditioned failure analysis, while deliberately excluding the agent's current state to avoid reintroducing stale claims after restoration.

Formally, we characterize RIR through a unified recovery operator parameterized by a
rollback depth $k$ and the memory $\mathcal{M}^{+}$ retained across the recovery boundary. We show that common correction and rollback mechanisms are restricted cases of the unified operator, and that their induced recovery-policy classes are therefore contained within the RIR recovery space. This policy-class inclusion leads directly to an optimal-value monotonicity result, under which the best task-completion probability attainable by RIR is no lower than that of any such restricted mechanism. Empirically, RIR consistently outperforms representative baselines across multiple long-horizon benchmarks and LLM backbones, improving average success rate by up to 6.57 percentage points while maintaining selective recovery under constrained interaction budgets. Our contributions are as follows:
\begin{itemize}
    \item We recast trajectory contamination in long-horizon agent tasks as a problem
    of \textit{rollback boundary} placement and identify three coupled questions that
    recovery must resolve: when to intervene, where to resume, and what information
    should survive the rollback.

    \item We propose \textbf{Rollback-Induced Reflection (RIR)}, which combines adaptive review, coarse-to-fine restore localization, and rollback-consistent reflection to recover execution state without discarding reusable knowledge from failed branches.

    \item We formalize RIR through a unified recovery operator
    and establish policy-class containment and optimal-value
    monotonicity for mechanisms that restrict rollback depth
    within a shared memory-update space. Experiments on long-horizon benchmarks further validate its effectiveness under constrained interaction budgets.
\end{itemize}
\section{Related Work}

LLM agents increasingly use context-management mechanisms such as external memory,
experience retrieval, and selective context construction to support long-horizon
interaction~\citep{yu2026agentic,chhikara2025mem0,jia2026agent,lu2026beyond}.
However, these methods generally do not address what happens once an erroneous action
has already altered the environment: whether and how execution itself should be
recovered~\citep{zhang2026calibration,zhang2026webrollback,hu2025webcot,wu2025backtrackagent}. We therefore focus on
rollback-oriented recovery and group prior work into two broad forms.

\textbf{Information-level correction.}
Self-Refine~\citep{madaan2023self}, AgenTracer~\citep{zhang2026agentracer},
Reflexion~\citep{shinn2023reflexion}, and ReflAct~\citep{kim2025reflact} improve
subsequent decisions through feedback or reflection, while ExpeL~\citep{zhao2024expel},
AutoGuide~\citep{fu2024autoguide}, and G-Memory~\citep{zhang2026g} reuse experience
across tasks. The effectiveness of such correction depends on the reliability of
the feedback~\citep{huang2024cannot,kamoi2024selfcorrection}. Moreover,
information-level correction alone does not restore environmental states altered
by earlier actions, and branch-local state claims retained in context may become
stale if execution is subsequently restored.

\textbf{State-level recovery.}
GA-Rollback~\citep{li2025generator} and
WebRollback~\citep{zhang2026webrollback} explicitly restore earlier states in
interactive trajectories, while BacktrackAgent~\citep{wu2025backtrackagent}
combines error detection with backtracking for GUI agents.
SRC~\citep{hao2026speculative} and DART~\citep{yang2026dart} study rollback
for training-data construction and structured recoverability, respectively.
GA-Rollback also evaluates a combination with Reflexion, demonstrating that
rollback and reflection can be used together.
Related search-based approaches, such as Tree of Thoughts~\citep{yao2023tree},
explore alternative reasoning paths, but reasoning-path backtracking does not
by itself specify how to restore an interactive environment or handle information
invalidated by that restoration. The closest concurrent work, AgentRewind~\citep{zhuang2026agentrewind}, checkpoints
aligned agent and environment states and preserves textual memory across rewinds, but
relies primarily on agent-initiated recovery and does not explicitly model the validity
of retained information after restoration.

In contrast, RIR focuses on the recovery boundary as an explicit control problem.
Its distinction is not the presence of rollback or reflection alone, but the
coordination of \emph{when} to review, \emph{where} to resume, and \emph{what}
information should survive recovery. RIR maintains one active execution trajectory
with selective recovery and separates branch-local state restored with the
checkpoint from reusable knowledge retained across attempts. This separation
explicitly addresses the risk of treating observations from an abandoned branch
as claims about the current restored state.
\section{Problem Formulation}

We consider a partially observable long-horizon interactive task
$\mathcal{E}=(\mathcal{S},\mathcal{A},\mathcal{O},P,\Omega,R_g)$,
where $\mathcal{S}$, $\mathcal{A}$, and $\mathcal{O}$ denote the state, action, and
observation spaces. The environment evolves according to
$P(s_{t+1}| s_t,a_t)$ and emits observations through
$\Omega(o_{t+1}| s_{t+1})$. Given task objective $g$, the outcome function
$R_g(s_T,y_T)$ evaluates the terminal execution between the final state $s_T$ and the ground truth state $y_T$, it is binary for verifiable tasks and
may be real-valued when graded rewards are available.

At step $t$, the base agent follows an LLM policy
$
a_t\sim\pi_\theta(\cdot| g,h_t,\mathcal{M}),
$
where $\mathcal{M}$ is the persistent
Reflection Memory and
$h_t=(o_0,a_0,\ldots,a_{t-1},o_t)$
is the interaction history of the current branch. The key distinction is that $h_t$ is checkpointed with execution,
whereas $\mathcal{M}$ is maintained outside checkpoints and may carry knowledge across
recovery attempts. The action space consists of final responses, ordinary task tools, and a recovery-control
tool:
$
\mathcal{A}
=
\mathcal{A}_{\mathrm{resp}}
\cup
\mathcal{A}_{\mathrm{tool}},
\mathcal{A}_{\mathrm{tool}}
=
\mathcal{A}_{\mathrm{task}}
\cup
\{\texttt{rollback}\}.
$
Actions in $\mathcal{A}_{\mathrm{task}}$ interact with the environment, while
$\mathcal{A}_{\mathrm{resp}}$ terminates the episode with a final response.
The \texttt{rollback} tool instead submits a recovery request and does not directly
modify the environment.

\textbf{Rollback.}
Before each executable action, the system stores a checkpoint
$
C_i=(s_i,h_i),
$
containing the environment state and branch-local context at step $i$.
Let
$
\mathcal{I}_t
=
\{\,i\le t\mid C_i \text{ is restorable at step } t\,\}
$
denote the set of admissible restore points. For a selected restore point $r\in\mathcal{I}_t$,
we define the trajectory discarded by recovery as the \emph{abandoned
suffix}
$
\tau_{r:t}
=
(a_r,o_{r+1},\ldots,a_{t-1},o_t).
$

A rollback first extracts reusable knowledge from $\tau_{r:t}$ to update the Reflection
Memory, and then restores the environment and branch-local context to $(s_r,h_r)$.
Execution subsequently resumes from the restored checkpoint under the updated memory.
We define
$
k=t-r
$
as the \emph{rollback depth}. 
We allow the limiting case $k=0$,
corresponding to context correction without state restoration.

\textbf{Objective.} RIR leaves the parameters $\theta$ of the base agent unchanged and instead introduces
a test-time recovery control policy $\Pi$. Given an agent-call budget
$B_{\mathrm{agent}}$ and a rollback budget $B_{\mathrm{rb}}$, RIR seeks to maximize
task success under bounded interaction:
\begin{equation}
\max_{\Pi}\;
\mathbb{E}_{\pi_\theta,\Pi,\mathcal{E}}
\!\left[R_g(s_T,y_T)\right]
\quad
\text{s.t.}
\quad
N_{\mathrm{agent}}\le B_{\mathrm{agent}},
\;\;
N_{\mathrm{rb}}\le B_{\mathrm{rb}}.
\label{eq:objective}
\end{equation}

\section{Rollback-Induced Reflection Framework}
\subsection{Overview}

\begin{figure}[t]
\begin{center}
\includegraphics[width=0.9\linewidth]{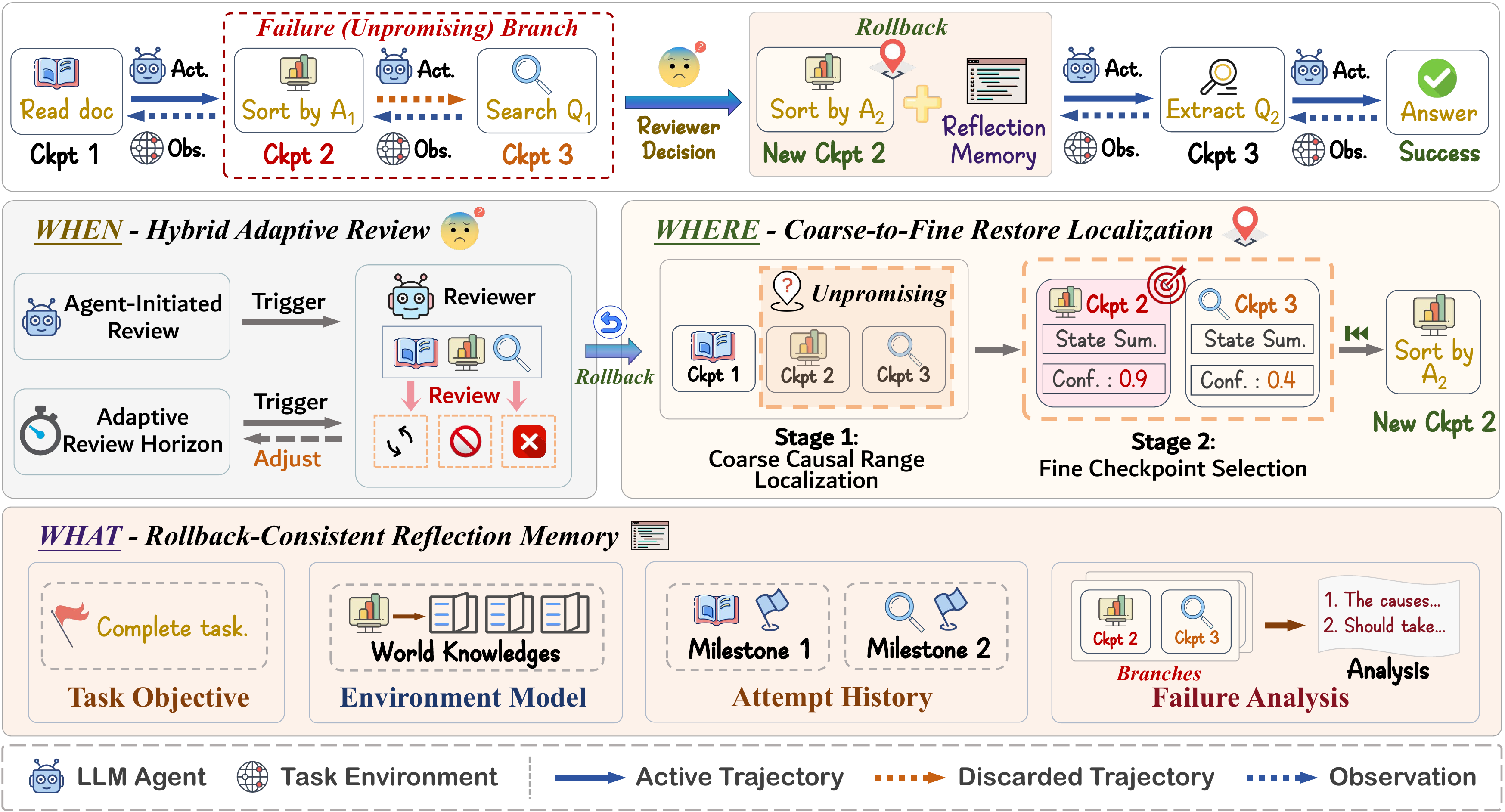}
\end{center}
\caption{Overview of the Rollback-Induced Reflection (RIR) framework.}
\label{fig:framework}
\vspace{-10pt}
\end{figure}

As shown in Figure~\ref{fig:framework}, RIR organizes recovery around three coupled
decisions: \textbf{When} decides whether the current branch should continue,
\textbf{Where} selects the checkpoint to resume from, and \textbf{What} determines
which information from the abandoned suffix should survive the rollback.
Algorithm~\ref{alg:rir} summarizes the complete execution loop.

During execution, RIR stores checkpoints along the current branch. A review is triggered
either by an agent recovery request or by the adaptive review schedule. If the branch is judged promising, the
system continues and adjusts the next review interval. Otherwise, RIR first localizes a restore checkpoint coarse-to-fine, then distills reflective knowledge from the suffix about to be abandoned. Finally, the system restores the environment and the agent context while retaining the updated Reflection Memory to guide the new attempt. The procedure can be summarized as:
\begin{equation}
\mathcal{M}^{+}=\phi(\mathcal{M},\tau_{r:t}),
\quad
(s_t,h_t,\mathcal{M})
\;\xrightarrow{\;\text{rollback}\;}\;
(s_{r},h_{r},\mathcal{M}^{+}),
\quad
r = t - k,
\label{eq:asymmetry}
\end{equation}
where $\phi$
denotes the information update instantiated in
Section~\ref{sec:what}. Together with rollback depth $k$, the pair
$(k,\mathcal{M}^+)$ defines the unified recovery operator analyzed in
Section~\ref{sec:theory}.
Equation~\ref{eq:asymmetry} makes RIR's central asymmetry explicit: the execution
world and branch-local context return to the past, while reflective knowledge acquired
from interactions that have already occurred is carried forward.

\subsection{When: Hybrid Adaptive Review}

Intervening too early truncates legitimate exploration, whereas intervening too late
allows errors to compound. To strike a balance, RIR invokes a single reviewer through two
complementary triggers: \textbf{agent-initiated review}, where the agent calls
\texttt{rollback} as a soft recovery request, and \textbf{adaptive scheduled review},
where the controller inspects the trajectory after a dynamic review horizon.
The former captures failures recognized by the agent itself, while the latter detects
stagnation or errors that the agent may overlook. Their combination reduces reliance
on either imperfect signal alone: agent requests enable timely intervention, while
scheduled reviews provide an external safeguard against unnoticed failure.
The rollback tool schema is provided in Appendix~\ref{app:rollback-tool}. 

Let $t_{\mathrm{rev}}$ denote the most recent review step and $L_t$ the current
review horizon. A review is triggered when:
\begin{equation}
\operatorname{ReviewEvent}_t
=\mathbb{I}\!\left[a_t=\texttt{rollback}\right]
\;\lor\;
\mathbb{I}\!\left[t-t_{\mathrm{rev}}\ge L_t\right].
\end{equation}
Importantly, calling \texttt{rollback} only requests review and does not directly
restore the environment. This separation prevents the agent's local uncertainty from
being converted immediately into an irreversible recovery decision. When a review is triggered, the LLM-based reviewer evaluates the task objective $g$,
current trajectory $h_t$, Reflection Memory $M$, and auxiliary execution signals $\xi_t$:
\begin{equation}
(\delta_t,\rho_t,f_t,\Delta E_t)
=
D_\psi(g,h_t,\mathcal{M},\xi_t),
\end{equation}
where $\delta_t\in\{\textsc{Continue},\textsc{Rollback}\}$ is the recovery verdict,
$\rho_t\in[0,1]$ controls the urgency of the next review, $f_t$ is the failure diagnosis
used when recovery is selected, and $\Delta E_t$ contains newly established environment
knowledge. Here, $\xi_t$ summarizes lightweight execution signals such as repeated
actions, no-effect steps, and state cycles. If the reviewer elects to continue, it uses $\rho_t$ to set the next review horizon:
\begin{equation}
\label{five}
L_{t+1}=L_{\min}+(1-\rho_t)\,(L_{\max}-L_{\min}),
\end{equation}
so higher urgency leads to earlier re-evaluation. This adaptive horizon concentrates
review effort on uncertain trajectories while allowing stable branches longer
uninterrupted exploration. More details and the instruction prompt of the reviewer are provided in
Appendix~\ref{app:review}.

\subsection{Where: Coarse-to-Fine Restore Localization}

Once recovery is approved, RIR must select a restore point from the stored checkpoints. Identifying the precise erroneous step in a single pass over the full trajectory is unreliable: failures in long-horizon tasks often arise from several interdependent decisions, while adjacent checkpoints may differ only marginally. We therefore adopt a two-stage coarse-to-fine localization procedure.

In the \textbf{coarse causal range localization} stage, an LLM-based selector uses the conditioned failure diagnosis $f_t$ together with the executed trajectory to identify a historical interval likely to contain the decisive error or an unmet precondition:
\begin{equation}
[l,u]=S_{\mathrm{coarse}}(g,h_t,\mathcal{M},f_t),
\quad l,u\in\mathcal{I}_t,\;\; l\le u.
\end{equation}
Rather than prematurely committing to a single culprit action, this stage eliminates large portions of the trajectory that are unlikely to have contributed to the current failure and restricts the subsequent search to a compact causal range.

In the \textbf{fine checkpoint selection} stage, the selector compares compact state
summaries $\sigma(C_i)$ within the localized interval and chooses a restore point that
removes the current obstruction while preserving as much valid progress as possible:
\begin{equation}
r=S_{\mathrm{fine}}\!\left(
g,f_t,\mathcal{M},
\{(i,\sigma(C_i))\mid i\in\mathcal{I}_t,\; l\le i\le u\}
\right).
\end{equation}
Here $\sigma(C_i)$ denotes a compact description derived from the checkpointed state,
such as completed sub-goals, current location, and key resources; it is computed from
the checkpoint and is not maintained as an additional memory variable. Let
$\operatorname{Escape}(\sigma(C_i),f_t)$ indicate whether restoring checkpoint $C_i$
removes the failure condition characterized by $f_t$. RIR follows the selection
principle:
\begin{equation}
r^{\star}
=
\max\Big\{
i\in[l,u]\cap\mathcal{I}_t
\;\Big|\;
\operatorname{Escape}(\sigma(C_i),f_t)=1
\Big\}.
\label{eq:latest-escape}
\end{equation}
In other words, RIR first filters out checkpoints whose restored states still retain
the obstruction identified in $f_t$, and then selects the \emph{latest} checkpoint
among the remaining candidates. The first criterion avoids immediately recreating the
same failure condition after restoration, while the second preserves the longest valid
prefix of the executed trajectory.

In practice, $S_{\mathrm{fine}}$ approximates
Equation~\ref{eq:latest-escape} by jointly comparing the candidate state summaries
against the conditioned failure diagnosis, and returns a restore point together with
its justification and confidence score. The coarse stage identifies \emph{which region
of the trajectory is causally relevant to the failure}, whereas the fine stage
determines \emph{which restorable state within that region provides the best point of
re-entry}. This coarse-to-fine decomposition avoids reducing restore localization to
single-step error attribution and instead balances two competing objectives: removing
the conditions that caused the failure and preserving as much correct progress as
possible. The detailed selector prompts are provided in Appendix~\ref{app:selector}.

\subsection{What: Rollback-Consistent Reflection Memory}
\label{sec:what}

State restoration determines which trajectory suffix is removed, but not what
information from that suffix should survive. Discarding it entirely loses reusable
experience, whereas retaining it verbatim may preserve branch-local state claims that
rollback has already invalidated. RIR therefore maintains a structured Reflection
Memory $\mathcal{M}$ outside the checkpoint:
\begin{equation}
\mathcal{M}=(G,E,H,F).
\end{equation}
\textbf{Task Objective ($G$).}
Stores a rollback-invariant representation of the final objective and provides stable
task-level guidance to the reviewer and the selector.

\textbf{Environment Model ($E$).}
Stores reusable knowledge about the environment across observed branches. We organize it into three semantic channels
$
E
=
E^{\mathrm{obs}}
\cup
E^{\mathrm{elim}}
\cup
E^{\mathrm{aff}},
$
corresponding respectively to observed facts, evidence-supported eliminations, and action or tool
affordances.
Entries are organized through constrained structured channels intended to reduce the risk of treating information from an abandoned branch as current state. Observations that depend on actions subsequently undone by rollback should be treated as historical evidence requiring re-verification, rather than as facts about the restored world.

\textbf{Attempt History ($H$).}
Records milestones and routes achieved in previous attempts as historical facts rather
than claims about the current state. A later attempt can therefore
reuse a discovered route without assuming that resources, locations, or progress
obtained before rollback remain valid after restoration. The currently executing
branch remains represented by $h_t$; its verified milestones are incorporated into
$H$ only when that attempt terminates or is abandoned.

\textbf{Failure Analysis ($F$).}
Summarizes why the most recently abandoned branch failed under the conditions that were in force at the time.
Failure patterns that become supported across attempts are instead promoted to the
Environment Model. Thus, $F$ explains \emph{why the latest branch failed under its
specific conditions}, whereas $E$ captures \emph{what is reusable about how the
environment behaves}.

Crucially, none of these fields represents the agent's current branch-local state;
that information is supplied by the active trajectory $h_t$ and the restored checkpoint
$(s_r,h_r)$. This separation prevents stale state claims from crossing the rollback
boundary. At each review, the reviewer extracts newly established environment knowledge
$\Delta E_t$. If rollback is executed, the reflection patcher additionally examines
the abandoned suffix $\tau_{r:t}$ to extract historical milestones $\Delta H_t$ and
construct a new failure analysis $F_{\mathrm{new}}$. The resulting update is:
\begin{equation}
E^{+}=E\cup\Delta E_t,
\quad
H^{+}=H\cup\Delta H_t,
\quad
F^{+}=F_{\mathrm{new}},
\quad
\mathcal{M}^{+}=(G,E^{+},H^{+},F^{+}).
\label{eq:update}
\end{equation}
Thus, $G$ remains fixed, $E$ and $H$ accumulate reusable knowledge and history, while
$F$ is replaced after each failed attempt. We denote this field-specific update by
$\phi$, which is applied in Equation~\ref{eq:asymmetry} to produce the updated memory $\mathcal{M}^{+}$.

The updated memory $\mathcal{M}^{+}$ replaces the previous Reflection Memory block in
the system prompt rather than being appended to it. Because $\mathcal{M}$ is maintained
outside checkpoints, restoring $C_r$ reverts the environment and branch-local context
while preserving the updated memory for the next attempt. This realizes the central
asymmetry of RIR: \textbf{rollback the world, keep the reflection}.

\begin{algorithm}[t]
\caption{\textbf{Rollback-Induced Reflection (RIR)}}
\label{alg:rir}

\begin{algorithmic}[1]

\Require Task objective $g$, base agent $\pi_\theta$,
rollback budget $B_{\mathrm{rb}}$
\State Initialize Reflection Memory $\mathcal{M}$,
review horizon $L\gets L_0$, and $N_{\mathrm{rb}}\gets0$

\While{task not terminated}

    \State Store checkpoint $C_t=(s_t,h_t)$
    \State Sample $a_t\sim\pi_\theta(\cdot\mid g,h_t,\mathcal{M})$

    \If{$a_t$ is a final response}
        \State \Return $a_t$
    \EndIf

    \If{$a_t=\texttt{rollback}$ \textbf{or} review horizon is reached}
        \WhenTag{review the current branch}

        \State Reviewer evaluation $(\delta_t,\rho_t,f_t,\Delta E_t)
        \gets \mathcal{D}_\psi(g,h_t,\mathcal{M},\xi_t)$
        \State Merge $\Delta E_t$ into the Environment Model
        
        \If{$\delta_t=\textsc{Rollback}$
        \textbf{and} $N_{\mathrm{rb}}<B_{\mathrm{rb}}$}

            \State $[l,u]\gets
            \mathcal{S}_{\mathrm{coarse}}(g,h_t,\mathcal{M},f_t)$
            \WhereTag{localize a causal restore range}

            \State Select restore point
            $r\gets\mathcal{S}_{\mathrm{fine}}
            (g,f_t,\mathcal{M},\{(i,\sigma(C_i))\; |\; l \leq i \leq u\})$

            \State Update
            $\mathcal{M}^{+}\gets
            \phi(\mathcal{M},\tau_{r:t})$
            \WhatTag{retain reusable knowledge}

            \State Restore the environment and branch-local context to $C_r$
            \State Reset the review horizon and increment $N_{\mathrm{rb}}$
            \State \textbf{continue}

        \Else
            \State Adapt the next review horizon using $\rho_t$
        \EndIf

    \EndIf

    \If{$a_t\neq\texttt{rollback}$}
        \State Execute $a_t$ and update the trajectory
    \EndIf

\EndWhile

\end{algorithmic}
\end{algorithm}

\subsection{Unified Recovery-Space Characterization}
\label{sec:theory}

Suppose recovery is triggered at step $t$, with restore point
$r=t-k\in\mathcal{I}_t$. Let
$\mathbb{M}(\mathcal{M},\tau_{r:t})$ denote the set of memory
configurations reachable by updating $\mathcal{M}$ from the
abandoned suffix $\tau_{r:t}$, including the unchanged memory
$\mathcal{M}$ and the full update
$\phi(\mathcal{M},\tau_{r:t})$. The unified recovery operator is
\begin{equation}
\mathcal{R}_{k,\mathcal{M}^{+}}:
(s_t,h_t,\mathcal{M})
\mapsto
(s_r,h_r,\mathcal{M}^{+}),
\quad
k\in\mathcal{K}_t,\;
\mathcal{M}^{+}\in\mathbb{M}(\mathcal{M},\tau_{r:t}).
\label{eq:unified_recovery_operator}
\end{equation}
RIR instantiates this operator through its restore selector
and memory updater $\phi$, yielding
Equation~\ref{eq:asymmetry}.
Since the abandoned suffix depends on the rollback depth,
the joint recovery space is
$\mathcal{C}_t
=
\{(k,\mathcal{M}^{+})
\mid k\in\mathcal{K}_t,\;
\mathcal{M}^{+}\in
\mathbb{M}(\mathcal{M},\tau_{t-k:t})\}$.

\textbf{Proposition 1 (Representational Containment).}
Consider a recovery mechanism $j$ that shares the same
admissible memory configurations at each rollback depth
but limits its depth choices to
$\mathcal{K}_{j,t}\subseteq\mathcal{K}_t$.
Its configuration space is therefore
$S_{j,t}
=
\{(k,\mathcal{M}^{+})\in\mathcal{C}_t
\mid k\in\mathcal{K}_{j,t}\}
\subseteq\mathcal{C}_t$.
This restriction affects only the available rollback depths;
memory retention remains unrestricted within the shared
memory family.
Examples include restart-based recovery with $k=t$,
fixed-depth rollback, and bounded-depth rollback,
whenever the corresponding depths are admissible.

Let $\Pi_{\mathrm{RIR}}$ denote the idealized policy class
with access to all configurations in $\mathcal{C}_t$
at each recovery event, and let $\Pi_j$ denote the class
restricted to $S_{j,t}$.
With identical base agents, available information, recovery
opportunities, and budget-accounting rules, the unrestricted
class can reproduce every decision available to the
restricted class. Thus,
$\Pi_j\subseteq\Pi_{\mathrm{RIR}}$.

This inclusion also orders the optimal values of the two
policy classes. For verifiable tasks, let
$R_g\in\{0,1\}$ indicate task success,
$J(\pi)=\Pr_{\pi}(R_g=1)$ denote the task-completion
probability of policy $\pi$, and
$V^\star(\Pi)=\sup_{\pi\in\Pi}J(\pi)$ denote the
optimal value over a policy class $\Pi$.

\textbf{Corollary 1 (Optimal-Value Monotonicity).}
For a restricted mechanism satisfying Proposition~1,
\begin{equation}
V^\star(\Pi_{\mathrm{RIR}})
\ge
V^\star(\Pi_j).
\label{eq:recovery_value_monotonicity}
\end{equation}
The inequality follows by taking the supremum over the
nested policy classes and also holds for expected return.
Full statements are provided in
Appendix~\ref{app:theory}.
\section{Experiments}


\subsection{Experimental Setup}
\label{sec:experimental_setup}

\textbf{Datasets \& metrics.}
We evaluate RIR on three long-horizon agent benchmarks:
ALFWorld~\citep{shridhar2020alfworld},
ScienceWorld~\citep{wang2022scienceworld}, and
GAIA~\citep{mialon2024gaia}, covering embodied interaction,
scientific experimentation, and open-domain tool use, respectively.
We report success rate (SR) on ALFWorld and ScienceWorld, and use an
LLM-as-a-Judge evaluator on GAIA to assess semantic equivalence between
predicted and reference answers. On ScienceWorld, we additionally report
normalized dense reward (DR) to measure partial task progress.

\textbf{Baselines.}
We compare against representative training-free test-time methods under the same
LLM backbones and interaction budgets:
ReAct~\citep{yao2022react} without explicit recovery;
Self-Refine~\citep{madaan2023self} and
Reflexion~\citep{shinn2023reflexion} for information-level correction;
LATS~\citep{zhou2023language} for search-based recovery;
and GA-Rollback~\citep{li2025generator} for explicit state rollback.
We use Qwen3-14B~\citep{yang2025qwen3}, DeepSeek-V3~\citep{liu2024deepseek} and Qwen3.8-27B as backbone models and follow official
implementations and recommended configurations whenever available.
Further baseline details are provided in
Appendix~\ref{app:baseline_details}.

\textbf{Implementation details.}
We implement the base agent with AgentScope~\citep{gao2025agentscope}.
Within each backbone setting, the acting agent and all RIR components use the
same LLM, so recovery gains cannot be attributed to a stronger auxiliary model.
All methods share identical task interfaces and interaction budgets. For GAIA, we additionally restore the workspace state during rollback, including
files created or modified by the agent, to keep the external environment consistent
with the recovered trajectory. Separately, final-answer evaluation on GAIA uses an
independent Qwen3.8-Max judge that does not participate in trajectory generation
or recovery.
Additional implementation details are provided in
Appendix~\ref{app:implementation_details}.


\subsection{Main Results}
\label{sec:main_results}

\textbf{Overall performance.}
As shown in Table~\ref{tab:main_results}, RIR achieves the highest average SR under all three backbones and the highest SR in eight of the nine backbone--benchmark settings. With Qwen3-14B, DeepSeek-V3, and Qwen3.8-27B, RIR reaches average success rates of 47.00\%, 69.43\%, and 76.54\%, outperforming the strongest corresponding baseline by 3.09, 6.57, and 3.04 percentage points, respectively. RIR also obtains the highest ScienceWorld DR under all three backbones. The only exception is GAIA with Qwen3.8-27B, where Reflexion leads by a narrow margin of 1.97 percentage points. We attribute this in part to the experimental setting of a 24-step interaction budget, which leaves limited room for RIR to realize the benefits of rollback through post-recovery exploration. Despite this isolated case, RIR's consistent average gains across backbones of varying capability suggest that explicit recovery remains useful rather than being subsumed by improvements in base-model capability. This empirical pattern is consistent with our recovery-space analysis: the broader recovery space provides additional useful choices in practice, although the theoretical result itself is a capacity statement rather than a performance guarantee.

\begin{table}[t]
\centering
\vspace{-10pt}
\caption{
Main results across three benchmarks.
SR denotes success rate and DR denotes normalized dense reward;
Avg.\ SR is the unweighted mean across the three benchmarks.
\textbf{Bold} and \underline{underlined} denote the best and second-best
results within each backbone, respectively.
}
\label{tab:main_results}

\footnotesize
\setlength{\tabcolsep}{9pt}
\renewcommand{\arraystretch}{0.85}

\begin{tabular}{@{}ll cccc c@{}}
\toprule

\multirow{2}{*}{\textbf{Backbone}}
&
\multirow{2}{*}{\textbf{Method}}
&
\textbf{ALFWorld}
&
\multicolumn{2}{c}{\textbf{ScienceWorld}}
&
\textbf{GAIA}
&
\multirow{2}{*}{\textbf{Avg.\ SR (\%)}} \\

\cmidrule(lr){3-3}
\cmidrule(lr){4-5}
\cmidrule(lr){6-6}

&
&
SR(\%) $\uparrow$
&
SR(\%) $\uparrow$
&
DR $\uparrow$
&
SR(\%) $\uparrow$
& \\

\midrule

\multirow{6}{*}{Qwen3-14B}

& ReAct
& 67.16
& 31.37
& 0.412
& 19.08
& 39.20 \\

& Reflexion
& 74.63
& \underline{35.38}
& \underline{0.426}
& \underline{21.71}
& \underline{43.91} \\

& LATS
& 67.91
& 29.15
& 0.407
& 15.79
& 37.62 \\

& Self-Refine
& 66.42
& 21.03
& 0.402
& 15.13
& 34.19 \\

& GA-Rollback
& \underline{77.61}
& 26.94
& 0.387
& 18.42
& 40.99 \\

\rowcolor{rirhi}
& \textbf{RIR (Ours)}
& \textbf{81.30}
& \textbf{36.67}
& \textbf{0.448}
& \textbf{23.03}
& \textbf{47.00} \\

\midrule

\multirow{6}{*}{DeepSeek-V3}

& ReAct
& \underline{85.82}
& \underline{67.89}
& \underline{0.779}
& \underline{34.87}
& \underline{62.86} \\

& Reflexion
& 77.61
& 62.36
& 0.735
& 34.21
& 58.06 \\

& LATS
& 61.65
& 65.31
& 0.758
& 32.89
& 53.28 \\

& Self-Refine
& 80.60
& 51.66
& 0.535
& \underline{34.87}
& 55.71 \\

& GA-Rollback
& 62.69
& 53.51
& 0.639
& 26.97
& 47.72 \\

\rowcolor{rirhi}
& \textbf{RIR (Ours)}
& \textbf{94.03}
& \textbf{76.75}
& \textbf{0.813}
& \textbf{37.50}
& \textbf{69.43} \\

\midrule

\multirow{6}{*}{Qwen3.8-27B}

& ReAct
& 96.27
& 70.11
& 0.813
& 49.34
& 71.91 \\

& Reflexion
& \underline{97.01}
& 70.85
& 0.806
& \textbf{52.63}
& \underline{73.50} \\

& LATS
& 90.30
& 64.21
& 0.769
& 47.37
& 67.29 \\

& Self-Refine
& 91.04
& \underline{74.91}
& \underline{0.832}
& 48.68
& 71.54 \\

& GA-Rollback
& 89.55
& 72.32
& 0.786
& 48.03
& 69.97 \\

\rowcolor{rirhi}
& \textbf{RIR (Ours)}
& \textbf{99.25}
& \textbf{79.70}
& \textbf{0.872}
& \underline{50.66}
& \textbf{76.54} \\

\bottomrule
\end{tabular}

\vspace{-9pt}
\end{table}

\begin{wrapfigure}[20]{R}{0.49\textwidth}
    \centering
    \vspace{-10pt}

    \includegraphics[width=0.87\linewidth]
    {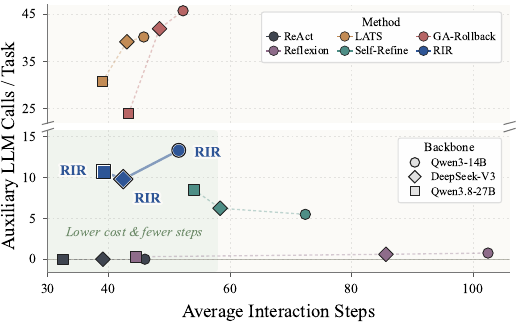}

    \vspace{-8pt}

    \caption{
    Efficiency trade-off on ScienceWorld across all three backbones.
    }
    \label{fig:scienceworld_efficiency}

    \vspace{2pt}

    \begingroup
    \footnotesize
    \setlength{\tabcolsep}{4pt}
    \renewcommand{\arraystretch}{0.95}

    \captionof{table}{
    Average executed rollbacks per task on ScienceWorld.
    Methods without state-level rollback are omitted.
    }
    \label{tab:scienceworld_rollback}

    \vspace{2pt}

    \begin{tabular}{@{}lccc@{}}
        \toprule
        \textbf{Method}
        & \textbf{Q.3-14B}
        & \textbf{D.S.-V3}
        & \textbf{Q.3.8-27B} \\
        \midrule
        GA-Rollback
        & 2.908
        & 2.910
        & 3.112 \\
        \textbf{RIR}
        & \textbf{0.978}
        & \textbf{0.920}
        & \textbf{1.266} \\
        \bottomrule
    \end{tabular}
    \endgroup

    \vspace{-8pt}
\end{wrapfigure}


\textbf{Efficiency.}
Figure~\ref{fig:scienceworld_efficiency} compares test-time overhead on
ScienceWorld across three backbones. RIR remains in the low-overhead regime while
requiring substantially fewer auxiliary LLM calls than the more expensive
recovery baselines, and this pattern is consistent across Qwen3-14B, DeepSeek-V3 and Qwen3.8-27B. Recovery is also selective rather than frequent:
Table~\ref{tab:scienceworld_rollback} shows fewer than 1.3 executed rollback per
task for RIR under all backbones, compared with approximately 3.0 for
GA-Rollback. Thus, the performance gains are associated with targeted correction
rather than repeated rollback and re-exploration.
Complete efficiency results are provided in
Appendix~\ref{app:efficiency_details}.

To complement the aggregate results, we further inspect representative successful
and unsuccessful recovery trajectories in Appendix~\ref{app:case_studies}.
The cases illustrate both when rollback provides state-level benefits that
reflection alone cannot recover and when incorrect diagnosis can cause recovery
to revisit an ineffective branch.


\subsection{Ablation Studies}
\label{sec:ablation}

\begin{figure}[t]
    \centering

    \begin{minipage}[t]{0.59\textwidth}
        \vspace{0pt}
        \centering

        \captionof{table}{
        Ablations on ScienceWorld with DeepSeek-V3 as the backbone.
        \textbf{Avg.\ Steps}, \textbf{Rev.}, \textbf{RB}, and
        \textbf{Aux.} report the per-task averages of interaction
        steps, reviewer invocations, executed rollbacks, and
        auxiliary LLM calls, respectively.
        }
        \label{tab:ablation_summary}

        \footnotesize
        \setlength{\tabcolsep}{3pt}
        \renewcommand{\arraystretch}{1.05}

        \resizebox{\linewidth}{!}{%
        \begin{tabular}{@{}lcccccc@{}}
            \toprule
            \textbf{Variant}
            & \textbf{SR$\uparrow$}
            & \textbf{DR$\uparrow$}
            & \textbf{Avg.\ Steps}
            & \textbf{Rev.}
            & \textbf{RB}
            & \textbf{Aux.} \\
            \midrule

            \rowcolor{black!4}
            \multicolumn{7}{c}{
                \textit{Review-trigger variants}
            } \\

            Agent-only
            & 73.43 & 0.800 & 44.2 & 3.68 & 0.86 & 6.28 \\

            Sched.-only (fix.)
            & 74.54 & 0.795 & 39.2 & 6.05 & 0.51 & 7.58 \\

            Sched.-only (adap.)
            & 72.32 & 0.788 & 39.0 & 6.13 & 0.54 & 7.76 \\

            Hybrid (fixed)
            & 69.00 & 0.761 & 46.6 & 7.79 & 1.17 & 11.32 \\

            \addlinespace[2pt]

            \rowcolor{black!4}
            \multicolumn{7}{c}{
                \textit{Recovery-component variants}
            } \\

            Reflection only
            & 70.48 & 0.738 & 36.5 & 7.25 & / & 8.12 \\

            Rollback only
            & 69.37 & 0.801 & 46.7 & 7.81 & 1.30 & 11.82 \\

            \midrule

            \rowcolor{rirhi}
            \textbf{RIR}
            & \textbf{76.75}
            & \textbf{0.813}
            & 42.4 & 7.08 & 0.92 & 9.85 \\

            \bottomrule
        \end{tabular}%
        }
    \end{minipage}%
    \hfill
    \begin{minipage}[t]{0.39\textwidth}
        \vspace{0pt}
        \centering

        \includegraphics[width=0.9\linewidth]
        {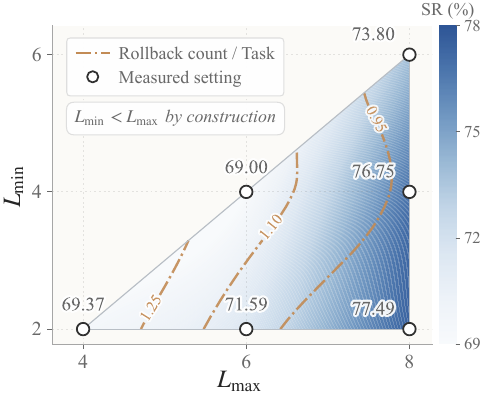}
        \vspace{-5pt}
        \captionof{figure}{
        Sensitivity to the adaptive review range
        $(L_{\min},L_{\max})$ on ScienceWorld with DeepSeek-V3.
        Color denotes SR and orange contours denote
        average rollback frequency.
        }
        \label{fig:horizon_ablation}
    \end{minipage}
\end{figure}

\textbf{When to review.}
We compare agent-only, scheduled-only, and hybrid review under fixed or adaptive
scheduling. As shown in Table~\ref{tab:ablation_summary}, the full
hybrid-adaptive RIR achieves the highest task performance
(76.75\% SR, 0.813 DR). Importantly, Hybrid (fixed) performs worse despite
triggering more reviews and rollbacks and using more auxiliary LLM calls.
Adaptive scheduling therefore appears to improve \emph{when} the two review
signals intervene, rather than simply increasing intervention frequency.
This supports the design of review timing as an adaptive control decision rather
than a fixed periodic mechanism.


\begin{figure}[t]
    \centering
    \includegraphics[width=0.9\linewidth]{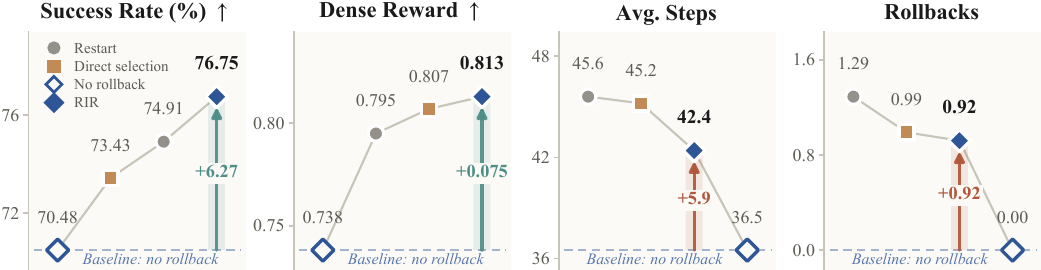}
    \caption{
    Where-to-recover ablation on ScienceWorld with DeepSeek-V3.
    \emph{No rollback} retains reflection but disables state restoration;
    \emph{Restart} restores to the initial checkpoint while retaining the same review policy and reflection mechanism as RIR;
    \emph{Direct selection} chooses a checkpoint in one stage;
    and \emph{RIR} uses coarse-to-fine restore localization.
    }
    \label{fig:where_ablation}
    \vspace{-15pt}
\end{figure}


\textbf{Where to recover.}
We next vary restore-point localization while keeping the review policy and
Reflection Memory unchanged. We compare \emph{No rollback} (reflection only),
\emph{Restart}, \emph{Direct selection}, and the full coarse-to-fine RIR
selector. Figure~\ref{fig:where_ablation} shows that No rollback produces the
shortest trajectories but substantially lower task performance, indicating that
information-level correction alone cannot replace state recovery. Among
rollback-based variants, RIR achieves the highest SR (76.75\%) and DR (0.813)
while requiring the fewest steps (42.4) and rollbacks (0.92).








\textbf{Rollback vs. Reflection.}
We further isolate the two core recovery components by comparing
\emph{Reflection only}, \emph{Rollback only}, and the full RIR.
Table~\ref{tab:ablation_summary} shows that neither component alone matches the
complete method: RIR improves SR by 6.27 points over Reflection only and by
7.38 points over Rollback only.  Reflection only is cheaper but cannot repair execution-state
errors, whereas Rollback only incurs greater interaction and inference cost
without preserving reusable knowledge across attempts.
In terms of our recovery operator, these variants separately restrict either
state restoration or the memory update, while RIR couples both dimensions.
The resulting gap therefore empirically supports their complementarity.


\textbf{Dynamic range setting.}
We finally vary the adaptive review range $(L_{\min},L_{\max})$.
Figure~\ref{fig:horizon_ablation} shows that the review range affects performance. An excessively large $L_{\min}$ delays scheduled reassessment, whereas an excessively small $L_{\max}$ restricts uninterrupted exploration. The best measured setting, $(2,8)$, achieves 77.49\% SR; the default $(4,8)$ achieves 76.75\%, a difference of 0.74 percentage points. Thus, the default remains competitive without selecting the best measured configuration. Under Equation~\ref{five}, $\rho$ controls the next review interval rather than directly deciding whether to roll back. The range bounds this adjustment, permitting earlier reassessment of high-urgency branches while preserving an exploration window. Together with Table~\ref{tab:ablation_summary}, where hybrid adaptive scheduling improves SR while reducing reviews, rollbacks, and auxiliary calls relative to hybrid fixed scheduling, these results are consistent with improved intervention timing rather than increased intervention frequency.
\section{Conclusion}

In this work, we propose Rollback-Induced Reflection (RIR), a unified recovery framework for long-horizon LLM agents. RIR couples adaptive review, restore localization, and persistent reflection to address a key limitation of existing recovery methods: correcting execution state without losing useful experience from failed trajectories. We further characterize recovery through a unified operator over rollback depth and updated reflection memory, showing that several common correction and rollback mechanisms arise as restricted cases of the RIR recovery space and establishing the corresponding optimal-value monotonicity result. Empirically, RIR improves task completion without relying on frequent intervention, while the ablation results show that review timing, restore localization, and reflection play complementary roles in successful recovery. More broadly, our results suggest that reliable long-horizon agents require recovery mechanisms that do more than undo mistakes: they must transform failed interaction into a better basis for subsequent decision-making.

\bibliography{iclr2027_conference}

@article{yao2023tree,
  title={Tree of thoughts: Deliberate problem solving with large language models},
  author={Yao, Shunyu and Yu, Dian and Zhao, Jeffrey and Shafran, Izhak and Griffiths, Tom and Cao, Yuan and Narasimhan, Karthik},
  journal={Advances in neural information processing systems},
  volume={36},
  pages={11809--11822},
  year={2023}
}

@article{zhang2026calibration,
  title={Calibration Is Not Control: Why LLM-Agent Oversight Needs Intervention},
  author={Zhang, Chubin and Wan, Zhenglin and Yu, Xingrui and Wu, Jingxuan and Wen, Qi and Zhou, Pengfei and Zhao, Wangbo and Tsang, Ivor},
  journal={arXiv preprint arXiv:2606.21399},
  year={2026}
}

@article{li2025hiplan,
  title={Hiplan: Hierarchical planning for llm-based agents with adaptive global-local guidance},
  author={Li, Ziyue and Chang, Yuan and Yu, Gaihong and Le, Xiaoqiu},
  journal={arXiv preprint arXiv:2508.19076},
  year={2025}
}

@article{yang2025qwen3,
  title={Qwen3 technical report},
  author={Yang, An and Li, Anfeng and Yang, Baosong and Zhang, Beichen and Hui, Binyuan and Zheng, Bo and Yu, Bowen and Gao, Chang and Huang, Chengen and Lv, Chenxu and others},
  journal={arXiv preprint arXiv:2505.09388},
  year={2025}
}

@inproceedings{park2023generative,
  title={Generative agents: Interactive simulacra of human behavior},
  author={Park, Joon Sung and O'Brien, Joseph and Cai, Carrie Jun and Morris, Meredith Ringel and Liang, Percy and Bernstein, Michael S},
  booktitle={Proceedings of the 36th Annual ACM Symposium on User Interface Software and Technology},
  pages={1--22},
  year={2023}
}

@article{liu2024deepseek,
  title={Deepseek-v3 technical report},
  author={Liu, Aixin and Feng, Bei and Xue, Bing and Wang, Bingxuan and Wu, Bochao and Lu, Chengda and Zhao, Chenggang and Deng, Chengqi and Zhang, Chenyu and Ruan, Chong and others},
  journal={arXiv preprint arXiv:2412.19437},
  year={2024}
}

@inproceedings{wei2022chain,
  title={Chain-of-thought prompting elicits reasoning in large language models},
  author={Wei, Jason and Wang, Xuezhi and Schuurmans, Dale and Bosma, Maarten and Xia, Fei and Chi, Ed and Le, Quoc V and Zhou, Denny and others},
  booktitle={Advances in Neural Information Processing Systems},
  volume={35},
  pages={24824--24837},
  year={2022}
}

@article{yao2022react,
  title={React: Synergizing reasoning and acting in language models},
  author={Yao, Shunyu and Zhao, Jeffrey and Yu, Dian and Du, Nan and Shafran, Izhak and Narasimhan, Karthik and Cao, Yuan},
  journal={arXiv preprint arXiv:2210.03629},
  year={2022}
}

@inproceedings{madaan2023self,
  title={Self-refine: Iterative refinement with self-feedback},
  author={Madaan, Aman and Tandon, Niket and Gupta, Prakhar and Hallinan, Skyler and Gao, Luyu and Wiegreffe, Sarah and Alon, Uri and Dziri, Nouha and Prabhumoye, Shrimai and Yang, Yiming and others},
  booktitle={Advances in Neural Information Processing Systems},
  volume={36},
  pages={46534--46594},
  year={2023}
}

@inproceedings{shinn2023reflexion,
  title={Reflexion: Language agents with verbal reinforcement learning},
  author={Shinn, Noah and Cassano, Federico and Gopinath, Ashwin and Narasimhan, Karthik and Yao, Shunyu},
  booktitle={Advances in Neural Information Processing Systems},
  volume={36},
  pages={8634--8652},
  year={2023}
}

@inproceedings{huang2024cannot,
  title={Large language models cannot self-correct reasoning yet},
  author={Huang, Jie and Chen, Xinyun and Mishra, Swaroop and Zheng, Huaixiu Steven and Yu, Adams and Song, Xinying and Zhou, Denny},
  booktitle={International Conference on Learning Representations},
  volume={2024},
  pages={32808--32824},
  year={2024}
}

@article{kamoi2024selfcorrection,
  title={When can LLMs actually correct their own mistakes? a critical survey of self-correction of llms},
  author={Kamoi, Ryo and Zhang, Yusen and Zhang, Nan and Han, Jiawei and Zhang, Rui},
  journal={Transactions of the Association for Computational Linguistics},
  volume={12},
  pages={1417--1440},
  year={2024}
}

@inproceedings{zhao2024expel,
  title={Expel: Llm agents are experiential learners},
  author={Zhao, Andrew and Huang, Daniel and Xu, Quentin and Lin, Matthieu and Liu, Yong-Jin and Huang, Gao},
  booktitle={Proceedings of the AAAI Conference on Artificial Intelligence},
  volume={38},
  pages={19632--19642},
  year={2024}
}

@inproceedings{fu2024autoguide,
  title={Autoguide: Automated generation and selection of context-aware guidelines for large language model agents},
  author={Fu, Yao and Kim, Dong-Ki and Kim, Jaekyeom and Sohn, Sungryull and Logeswaran, Lajanugen and Bae, Kyunghoon and Lee, Honglak},
  booktitle={Advances in Neural Information Processing Systems},
  volume={37},
  pages={119919--119948},
  year={2024}
}

@inproceedings{zhang2026g,
  title={G-memory: Tracing hierarchical memory for multi-agent systems},
  author={Zhang, Guibin and Fu, Muxin and Wang, Kun and Wan, Frank and Yu, Miao and Yan, Shuicheng},
  booktitle={Advances in Neural Information Processing Systems},
  volume={38},
  pages={12988--13018},
  year={2026}
}

@inproceedings{kim2025reflact,
  title={Reflact: World-grounded decision making in LLM agents via goal-state reflection},
  author={Kim, Jeonghye and Rhee, Sojeong and Kim, Minbeom and Kim, Dohyung and Lee, Sangmook and Sung, Youngchul and Jung, Kyomin},
  booktitle={Proceedings of the 2025 Conference on Empirical Methods in Natural Language Processing},
  pages={33421--33453},
  year={2025}
}

@article{zhou2023language,
  title={Language agent tree search unifies reasoning acting and planning in language models},
  author={Zhou, Andy and Yan, Kai and Shlapentokh-Rothman, Michal and Wang, Haohan and Wang, Yu-Xiong},
  journal={arXiv preprint arXiv:2310.04406},
  year={2023}
}

@inproceedings{li2025generator,
  title={Generator-assistant stepwise rollback framework for large language model agent},
  author={Li, Xingzuo and Chen, Kehai and Long, Yunfei and Bai, Xuefeng and Xu, Yong and Zhang, Min},
  booktitle={Proceedings of the 2025 Conference on Empirical Methods in Natural Language Processing},
  pages={17694--17711},
  year={2025}
}

@inproceedings{wu2025backtrackagent,
  title={Backtrackagent: Enhancing gui agent with error detection and backtracking mechanism},
  author={Wu, Qinzhuo and Gao, Pengzhi and Liu, Wei and Luan, Jian},
  booktitle={Proceedings of the 2025 Conference on Empirical Methods in Natural Language Processing},
  pages={4250--4272},
  year={2025}
}

@inproceedings{zhang2026webrollback,
  title={WebRollback: Enhancing Web Agents with Explicit Rollback Mechanisms},
  author={Zhang, Zhisong and Fang, Tianqing and Ma, Kaixin and Yu, Wenhao and Zhang, Hongming and Mi, Haitao and Yu, Dong},
  booktitle={Proceedings of the 19th Conference of the European Chapter of the Association for Computational Linguistics},
  pages={187--197},
  year={2026}
}

@article{hao2026speculative,
  title={Speculative Rollback Correction for Quality-Diverse Web Agent Imitation},
  author={Hao, Longkun and Lin, Hongyu and Li, Hao and Yang, Zhichao and Hao, Haojie and Huang, Dongshuo and Yang, Haitao and Ge, Hongyu and Wu, Yanjun and Yin, Zi Hao and others},
  journal={arXiv preprint arXiv:2606.12485},
  year={2026}
}

@article{yang2026dart,
  title={DART: Semantic Recoverability for Structured Tool Agents},
  author={Yang, Ke and Li, Panpan and Wu, Zonghan and Xu, Kejin and Huang, Huaxi and Huang, Xiaoshui},
  journal={arXiv preprint arXiv:2605.23311},
  year={2026}
}

@article{zhuang2026agentrewind,
  title   = {AgentRewind: Recoverable Execution for Long-Horizon {LLM} Agents},
  author  = {Zhuang, Yu and Chen, Kefei and Duan, Yitong and Zheng, Shuxin and Li, Jian and Zhang, Xu-Yao},
  journal = {arXiv preprint arXiv:2608.14380},
  year    = {2026},
}

@inproceedings{yu2026agentic,
  title={Agentic memory: Learning unified long-term and short-term memory management for large language model agents},
  author={Yu, Yi and Yao, Liuyi and Xie, Yuexiang and Tan, Qingquan and Feng, Jiaqi and Li, Yaliang and Wu, Libing},
  booktitle={Proceedings of the 64th Annual Meeting of the Association for Computational Linguistics},
  pages={21457--21483},
  year={2026}
}

@article{chhikara2025mem0,
  title={Mem0: Building production-ready ai agents with scalable long-term memory},
  author={Chhikara, Prateek and Khant, Dev and Aryan, Saket and Singh, Taranjeet and Yadav, Deshraj},
  journal={arXiv preprint arXiv:2504.19413},
  year={2025}
}

@inproceedings{jia2026agent,
  title={Agent-enhanced heterogeneous graph RAG for academic question answering},
  author={Jia, Runsong and Wu, Mengjia and Ding, Ying and Lu, Jie and Zhang, Yi},
  booktitle={Proceedings of the ACM Web Conference},
  pages={8765--8768},
  year={2026}
}

@inproceedings{lu2026beyond,
  title={Beyond the Context Window: Scaling Agentic RL via End-to-end Optimized Context Compression},
  author={Lu, Miao and Sun, Weiwei and Du, Weihua and Ling, Zhan and Yao, Xuesong and Liu, Kang and Chen, Jiecao},
  booktitle={Proceedings of the 64th Annual Meeting of the Association for Computational Linguistics},
  pages={21074--21125},
  year={2026}
}

@inproceedings{hu2025webcot,
  title={WebCoT: Enhancing Web Agent Reasoning by Reconstructing Chain-of-Thought in Reflection, Branching, and Rollback.},
  author={Hu, Minda and Fang, Tianqing and Zhang, Jianshu and Ma, Jun-Yu and Zhang, Zhisong and Zhou, Jingyan and Zhang, Hongming and Mi, Haitao and Yu, Dong and King, Irwin},
  booktitle={Proceedings of the 2025 Conference on Empirical Methods in Natural Language Processing},
  pages={5155--5173},
  year={2025}
}

@inproceedings{zhang2026agentracer,
  title={Agentracer: Who is inducing failure in the llm agentic systems?},
  author={Zhang, Guibin and Wang, Junhao and Chen, Junjie and Zhou, Wangchunshu and Wang, Kun and Yan, Shuicheng},
  booktitle={International Conference on Learning Representations},
  volume={2026},
  pages={11377--11399},
  year={2026}
}

@article{shridhar2020alfworld,
  title={Alfworld: Aligning text and embodied environments for interactive learning},
  author={Shridhar, Mohit and Yuan, Xingdi and C{\^o}t{\'e}, Marc-Alexandre and Bisk, Yonatan and Trischler, Adam and Hausknecht, Matthew},
  journal={arXiv preprint arXiv:2010.03768},
  year={2020}
}

@inproceedings{wang2022scienceworld,
  title={Scienceworld: Is your agent smarter than a 5th grader?},
  author={Wang, Ruoyao and Jansen, Peter and C{\^o}t{\'e}, Marc-Alexandre and Ammanabrolu, Prithviraj},
  booktitle={Proceedings of the 2022 Conference on Empirical Methods in Natural Language Processing},
  pages={11279--11298},
  year={2022}
}

@inproceedings{mialon2024gaia,
  title={Gaia: a benchmark for general ai assistants},
  author={Mialon, Gr{\'e}goire and Fourrier, Cl{\'e}mentine and Wolf, Thomas and LeCun, Yann and Scialom, Thomas},
  booktitle={International Conference on Learning Representations},
  volume={2024},
  pages={9025--9049},
  year={2024}
}

@article{gao2025agentscope,
  title={AgentScope 1.0: A developer-centric framework for building agentic applications},
  author={Gao, Dawei and Li, Zitao and Xie, Yuexiang and Kuang, Weirui and Yao, Liuyi and Qian, Bingchen and Ma, Zhijian and Cui, Yue and Luo, Haohao and Li, Shen and others},
  journal={arXiv preprint arXiv:2508.16279},
  year={2025}
}
\bibliographystyle{iclr2027_conference}

\appendix
\section{Prompt Design}
\label{a}
\subsection{The Rollback Tool}
\label{app:rollback-tool}

RIR exposes exactly one control tool to the base agent. Its interface is deliberately minimal: a single free-text \texttt{reason} field, no target step, and no rollback depth. The agent reports that it is stuck and why; it does not choose where the trajectory is restored, which is decided by the reviewer and the selector. The tool is declared read-only, so invoking it cannot alter the task environment: a request costs one step and leaves
the world untouched. Figure~\ref{fig:rollback-tool} shows the schema the model receives and the accompanying prompt instruction.

\begin{figure}[t]
\begin{minipage}{\linewidth}
\begin{tcolorbox}[
  colback=white, colframe=black!55, boxrule=0.4pt, arc=1.5pt,
  left=3pt, right=3pt, top=2pt, bottom=2pt,
  fonttitle=\bfseries\footnotesize, coltitle=black, colbacktitle=black!8,
  title={\texttt{Rollback} tool schema}]
\begin{lstlisting}[style=rirjson]
{
  "name": "Rollback",
  "description":
    "Request that the controller roll back to an earlier checkpoint and revise the reflection before continuing.",
  "input_schema": {
    "type": "object",
    "properties": {
      "reason": {
        "type": "string",
        "description":
          "Why the current trajectory should be rolled back."
      }
    },
    "required": ["reason"]
  },
  "is_read_only": true
}
\end{lstlisting}
\end{tcolorbox}

\begin{tcolorbox}[
  colback=white, colframe=black!55, boxrule=0.4pt, arc=1.5pt,
  left=3pt, right=3pt, top=2pt, bottom=2pt,
  fonttitle=\bfseries\footnotesize, coltitle=black, colbacktitle=black!8,
  title={Rollback instruction in the system prompt}]
{\ttfamily\scriptsize
You also have a Rollback tool. Call it when the current approach is no longer
worth continuing --- for example when recent actions repeat without progress,
contradict earlier verified findings, or rest on an assumption you now believe
is wrong. Rolling back returns you to an earlier state carrying a lesson about
what went wrong, which is usually cheaper than patching a bad trajectory.
Always give a concrete reason when you call it.

\medskip
\textrm{\itshape Appended to every step:}\\
If the current path should be abandoned instead, call the Rollback tool with a
concrete reason.
}
\end{tcolorbox}
\end{minipage}
\caption{The \texttt{rollback} control tool, including the schema exposed to the
agent and the system-prompt instruction governing its use. The tool takes only a
reason and is read-only, so a request cannot directly modify the environment.}
\label{fig:rollback-tool}
\end{figure}

\textbf{Soft-request semantics.}
A request is a report, not a command, and the controller may decline it. The three possible outcomes are all returned to the agent as the tool result, so the agent always learns how its request was handled:

\begin{itemize}
\item \textbf{Granted.} The tool result reads ``Rollback request submitted; the
controller decides whether to rewind,'' and control passes to restore localization.
\item \textbf{Declined by the Reviewer.} The request opens a review with the event
type \textsc{RollbackRequest}. If the verdict is \textsc{Continue}, the agent
receives ``Rollback request rejected by the reviewer,'' followed by the reviewer's
reason, and resumes on the same branch.
\item \textbf{Declined before review.} If the current attempt has committed no
environment action since the last rollback, the request is rejected without
consulting the reviewer: restoring an attempt that changed nothing would reproduce
the same state. The agent is told ``this attempt has taken no action since the last
rollback, so rolling back cannot change anything.''
\end{itemize}

\textbf{Prompt layering.}
The rollback instruction is an addendum, appended to the system prompt only when the
rollback capability is enabled. An agent configured without recovery never receives
it and is never shown the tool, so the no-rollback baselines run on the same prompt as the base agent, with no
mention of a mechanism they cannot use.

\subsection{The Review Process}
\label{app:review}

A review is invoked either when the scheduled horizon is reached or when the agent requests a rollback. The event type is included in the prompt, so the reviewer knows whether it is answering a periodic inspection or an explicit recovery request. In addition to the task objective, the current trajectory, and the Reflection Memory, the prompt supplies the executor's reported state and a small set of run-history counters referenced by the rules below. Figure~\ref{fig:reviewer} reproduces the role and rule sections of the reviewer prompt.

\begin{figure}[ht]
\centering
\begin{minipage}[t]{\textwidth}
\begin{tcolorbox}[
  colback=white, colframe=black!55, boxrule=0.4pt, arc=1.5pt,
  left=3pt, right=3pt, top=2pt, bottom=2pt,
  fonttitle=\bfseries\footnotesize, coltitle=black, colbacktitle=black!8,
  title={Reviewer prompt}]
{\ttfamily\scriptsize\raggedright\parindent=0pt
\textless role\textgreater\par
You review an agent's current attempt at one task and decide whether it should
continue or be rolled back to an earlier point, and you extend the environment model
with what this review newly established. You never act in the world yourself. The
environment model is the only memory you write; attempt history and the failure
analysis belong to the rollback patcher.\par
\textless/role\textgreater\par
\smallskip
\textless rules\textgreater\par
Decide in this order; apply the first rule that fits.\par
\smallskip
\hangindent=1.2em 1. Evidence boundary. Only an executed action - a payload with its
returned observation - interacted with the world. Reasoning text, plans, and steps
that executed nothing interact with nothing; subjects named only in reasoning were
not visited, queried, or examined.\par
\smallskip
\hangindent=1.2em 2. Report only what is new. Write findings, eliminations and
affordances THIS review established and the environment model still lacks. Do not
restate entries the model already holds: the system tracks whether an attempt is
still producing knowledge, and restating old entries destroys that signal.
Establishing nothing new is a normal outcome - return empty lists.\par
\smallskip
\hangindent=1.2em 3. Precondition check - before blaming anything. When an executed
action repeats with no effect, the likeliest cause is an unmet precondition: the
agent's own position or state, a required prior step, or an unprepared target. Read
executor\_state: it reports what the agent's own state is, and a blocked action is
very often blocked by it. That is knowledge about the environment's behaviour, not
about the task: record the demonstrated rule as an affordance and CONTINUE so the
precondition can be satisfied. Only when the precondition hypothesis has been tested
and refuted may the repeat count as a futility signal.\par
\smallskip
\hangindent=1.2em 4. Futility - the rollback case. ROLLBACK when returning to an
earlier state is what unblocks the goal: (a) the same action still repeating with no
new observation after its preconditions were verified; (b) a cycle over
already-settled subjects; (c) actions contradicting the environment model; (d) a plan
that provably cannot reach the goal; (e) continued work after the needed evidence was
already established; (f) the attempt has stopped learning - it keeps acting while
establishing nothing the environment model lacked. Read the counters: a high
actions\_without\_effect or revisited\_state\_ratio is that state measured directly,
and it does not become acceptable merely because some candidate somewhere remains
unexamined.\par
\smallskip
{\leftskip=1.2em\noindent The test is whether coverage is still EXPANDING, not whether
the search is complete. A search is almost never complete, so treating incompleteness
as a reason to keep going means never judging an attempt at all. An attempt whose
recent actions returned outcomes the model already held is orbiting, however many
candidates remain in principle.\par}
\smallskip
\hangindent=1.2em 5. Rollback cannot fix a belief. A rollback rewinds decisions, not
knowledge - the environment model survives it. If the problem is that the agent
believes something the model contradicts, rolling back changes nothing; CONTINUE.\par
\smallskip
\hangindent=1.2em 6. Impossibility and completion audit. Before agreeing that
something is impossible, check the model: any goal-relevant positive finding, any
incomplete examination, or any plausible candidate with no entry refutes the claim -
unsupported impossibility warrants CONTINUE. Never declare the task complete:
completion is judged only by the environment. When everything looks settled, re-read
task\_objective and keep working on whatever required transformation, property, or
qualifier is not yet evidence-backed.\par
\smallskip
\hangindent=1.2em 7. Default: CONTINUE. A NEW observation is progress even before the
objective is met, and expanding coverage is normal work. Give a fresh attempt (low
actions\_since\_rollback) extra patience. This default does not survive rule 4(f):
``it might still find it'' is not a new observation.\par
\smallskip
review\_urgency is the only scheduling signal: 1 = review again soon, 0 = allow a long
stretch.\par
\textless/rules\textgreater\par
}
\end{tcolorbox}
\end{minipage}
\caption{The reviewer instruction in the system prompt.}
\label{fig:reviewer}
\end{figure}

\begin{figure}[t]
\centering
\begin{minipage}{\textwidth}
\begin{tcolorbox}[title={Stage 1: Coarse causal range localization},
  colback=white, colframe=black!55, boxrule=0.4pt, arc=1.5pt,
  left=3pt, right=3pt, top=2pt, bottom=2pt,
  fonttitle=\bfseries\footnotesize, coltitle=black, colbacktitle=black!8]
{\ttfamily\scriptsize\raggedright\parindent=0pt
\textless role\textgreater\par
You propose a NARROW interval of checkpoints worth restoring to, given a failed attempt:
from the earliest point that plausibly precedes the failure's root cause, to the latest
point that still escapes it. Checkpoint i is the state immediately before action i. You do
not pick the exact checkpoint yet. The task memory is read-only context.\par
\textless /role\textgreater\par
\medskip
\textless rules\textgreater\par
\hangindent=1.2em 1. Only an action payload with its returned observation shows what
interacted with the world; a reasoning-only step may waste compute, but the subjects in
its text were not acted on.\par
\hangindent=1.2em 2. The interval's end is the latest checkpoint that still precedes the
earliest committing step of the failing course - restoring later would keep the mistake.
The interval's start is the earliest checkpoint whose extra rewind still buys materially
different context - rewinding further only discards sound work.\par
\hangindent=1.2em 3. Keep it narrow. An interval spanning most of the trajectory carries
no information and will be trimmed by the system; if you cannot justify a narrow interval
from the cited steps, you have not located the root cause. State the step indices your
bounds rest on.\par
\hangindent=1.2em 4. Return an inclusive interval of checkpoint indices drawn from
valid\_checkpoint\_indices.\par
\textless /rules\textgreater\par
}
\end{tcolorbox}
\begin{tcolorbox}[title={Stage 2: Fine checkpoint selection},
  colback=white, colframe=black!55, boxrule=0.4pt, arc=1.5pt,
  left=3pt, right=3pt, top=2pt, bottom=2pt,
  fonttitle=\bfseries\footnotesize, coltitle=black, colbacktitle=black!8]
{\ttfamily\scriptsize\raggedright\parindent=0pt
\textless role\textgreater\par
You pick the exact checkpoint to restore, strictly from the validated candidate set
(already limited to the justified interval, minus proven dead-ends). Checkpoint i is the
state immediately before action i. The task memory is read-only context.\par
\textless /role\textgreater\par
\medskip
\textless rules\textgreater\par
\hangindent=1.2em 1. Restoring a checkpoint restores the executor's own state along with
the world. Each candidate reports that state. If the attempt was blocked by an unmet
precondition, a candidate whose state carries the same blocker will reproduce the same
failure - prefer one whose state does not.\par
\hangindent=1.2em 2. Among candidates that escape the blocker, prefer the latest: it
retains the most sound work. Choose an earlier one only when the trajectory shows the
mistake was committed before it.\par
\hangindent=1.2em 3. Each candidate reports prior\_rollbacks: how many times that exact
checkpoint was already restored in this task without solving it. Re-restoring one repeats
an intervention already known not to work; prefer a lower count when the counts differ.\par
\hangindent=1.2em 4. Return exactly one member of candidate\_checkpoints; never invent or
round an index.\par
\textless /rules\textgreater\par
}
\end{tcolorbox}
\end{minipage}
\caption{The two rollback target selection prompts. Stage~1
proposes the interval, Stage~2 chooses within it.}
\label{fig:selector}
\end{figure}

The reviewer outputs:
\begin{equation}
(\delta_t,\rho_t,f_t,\Delta E_t),
\end{equation}
where $\delta_t\in\{\textsc{Continue},\textsc{Rollback}\}$ is the recovery
decision, $\rho_t$ controls the next review horizon, $f_t$ summarizes the
current obstruction, and $\Delta E_t$ contains environment knowledge established
by the reviewed trajectory. These outputs serve different purposes:
$\delta_t$ determines whether recovery is entered, $\rho_t$ regulates when the
branch should next be inspected, and $(f_t,\Delta E_t)$ provide causal and
environmental evidence for subsequent recovery and execution.

\textbf{Evidence-grounded review.}
The reviewer is deliberately constrained to distinguish observed interaction
from the agent's own reasoning. Only executed actions and their returned
observations can establish environment knowledge; plans, hypotheses, or mentioned
objects alone do not. This prevents speculative reasoning from being promoted
into persistent Reflection Memory. The reviewer similarly records only information
that is not already represented in the Environment Model, so repeated reviews
do not accumulate paraphrases of previously established facts.

\textbf{Local obstruction versus branch failure.}
A central behavior of the reviewer is to distinguish a locally repairable action
failure from a trajectory-level failure that warrants recovery. An ineffective
action does not immediately imply rollback: it may simply reveal an unmet
precondition, such as an incorrect agent location, missing preparation step, or
invalid target state. In such cases, the reviewer favors \textsc{Continue} and
records the observed affordance so that the executor can repair the precondition
within the current branch. Rollback is reserved for cases where the trajectory
itself has become unproductive---for example, when the agent cycles over settled
states, repeatedly violates established environment knowledge, or follows a plan
whose earlier decisions prevent further progress.

This distinction is important because recovery is useful only when changing the
execution state can remove the diagnosed obstruction. If the problem is purely
informational---for example, the current belief conflicts with evidence already
stored in Reflection Memory---restoring an earlier state would not remove that
belief. The reviewer therefore favors continued execution in such cases and
allows the persistent memory to correct subsequent decisions. Conversely, when
the obstruction is tied to a committed state or earlier action choice, the
reviewer produces a rollback diagnosis $f_t$ that is subsequently used by the
restore-point selector.

\textbf{Review urgency.}
The recovery verdict and the review schedule are intentionally separated.
A branch may still be worth continuing while exhibiting uncertainty, repeated
friction, or weak progress. Such cases can receive \textsc{Continue} together
with high urgency, causing the controller to revisit the branch after a shorter
horizon. Productive trajectories instead receive lower urgency and are allowed
to proceed for longer without interruption. Thus, the reviewer need not make a
premature binary recovery decision merely to ensure that a risky branch is
examined again soon.

Finally, agent-triggered and scheduled reviews share the same semantic reviewer
but arise from complementary signals. An explicit rollback request exposes
failure recognized by the executor itself, whereas scheduled review can identify
stagnation or causal inconsistency that the acting agent has not recognized.
This division underlies the hybrid review mechanism used in RIR: the agent can
request immediate inspection when necessary, while adaptive scheduling provides
an independent opportunity to detect failures that would otherwise persist.

\begin{figure}[t]
\centering
\begin{minipage}[t]{\textwidth}
\begin{tcolorbox}[title={Reflection memory prompt},
  colback=white, colframe=black!55, boxrule=0.4pt, arc=1.5pt,
  left=3pt, right=3pt, top=2pt, bottom=2pt,
  fonttitle=\bfseries\footnotesize, coltitle=black, colbacktitle=black!8]
{\ttfamily\scriptsize\raggedright\parindent=0pt
\textless role\textgreater\par
You write the task memory after an attempt was rolled back: why it failed, which beliefs
it invalidated, and which subgoals it genuinely attained. What you write
is everything the next attempt will know about this one.\par
\textless /role\textgreater\par
\medskip
\textless rules\textgreater\par
Write in this order of care; apply every rule.\par
\hangindent=1.2em 1. Reasoning is a proposal, not an event. Only an action payload with its
returned observation establishes what was attempted, observed, or attained; do not turn
subjects mentioned only in reasoning into facts or attained subgoals.\par
\hangindent=1.2em 2. Condition every cause. Each failure cause names the condition under
which the thing failed - the acting state, position, or missing prerequisite in effect.
Read restored\_executor\_state: it reports the state the next attempt will actually start
from, and when a repeated action had no effect, that state is the prime suspect. An action
that failed under one condition remains available under others, and the next attempt obeys
your words verbatim: a bare ``X does not work'' bans X everywhere.\par
\hangindent=1.2em 3. Hard ban on global verdicts. No item may assert an unsupported global
existence or impossibility claim (``X does not exist anywhere'', ``the task cannot be
completed'', ``the search was exhaustive'', ``all candidates were checked''): a failed
search proves only that the examined candidates lacked the target, and candidates with no
entry always remain.\par
\hangindent=1.2em 4. Invalidated beliefs take only specifics. Only beliefs disproven by
contradictory evidence, a verified error, or a task-contract violation qualify - never a
behaviour category the task requires (such as exploration), never a claim that is merely
unproven. For source or access failures, invalidate only the failed path and keep the
candidate unverified.\par
\hangindent=1.2em 5. Attained subgoals are reusable history. Record only intermediate
results the abandoned attempt demonstrably reached, each with the means that reached it, so
the next attempt can reuse the route. State the outcome, not the motion: an attempt that
merely moved, looked, or failed attained nothing and gets no entry. Most abandoned attempts
attain nothing - returning an empty list is the normal case, not a failure to comply.\par
\hangindent=1.2em 6. Never claim the present. History says what was once reached, never
what holds now: the rollback undid the state, not the history. Do not write a subgoal that
would read as a description of the current situation.\par
\hangindent=1.2em 7. Replace, do not accumulate. The failure analysis is a complete
replacement of the previous one: keep still-valid items, drop obsolete or contradicted
ones, and never carry two conflicting explanations. Attempt history is the opposite -
append-only; earlier attempts' records are not yours to rewrite.\par
\hangindent=1.2em 8. Stay inside your boundary. You write causes, invalidated beliefs and
attained subgoals - never instructions (the next move is the generator's decision), never
world facts or behavioural rules (they belong to the environment model, which the reviewer
owns and which survives your patch).\par
\textless /rules\textgreater\par
}
\end{tcolorbox}
\end{minipage}
\caption{The prompt of Reflection Memory extraction. }
\label{fig:reflection}
\end{figure}

\subsection{Rollback Target Selection}
\label{app:selector}

Rollback target selection proceeds in two stages. Both selectors are prompted with the task objective, the failure diagnosis produced by the review that triggered recovery, the executed trajectory, and the Reflection Memory as read-only context. Figure~\ref{fig:selector} reproduces the role and rule sections of both selector prompts.

In the first stage, the coarse selector proposes an inclusive interval of candidate checkpoints. The interval is chosen to cover the portion of the trajectory most relevant to the diagnosed failure: its upper boundary should precede the decision that committed the agent to the failing course, while its lower boundary should avoid rewinding beyond the point at which additional rollback would provide materially different decision context. The selector is given the admissible checkpoint indices and, when applicable, a maximum interval width.

In the second stage, the fine selector chooses a single restore point from the admissible checkpoints within the localized interval after candidates known to correspond to dead ends have been excluded. In addition to the shared context, it receives the interval selected by the coarse stage together with the corresponding rationale, a compact description of the execution state restored by each candidate checkpoint, and the number of times each checkpoint has previously been restored during the current task.

\subsection{Reflection Memory Extraction}
\label{app:reflection}

Reflection Memory is updated once per rollback, after the restore point has been selected and before checkpoint restoration is executed. The reflection patcher receives the abandoned suffix, the failure diagnosis produced by the review that triggered recovery, the current Reflection Memory, the index of the abandoned attempt, and the executor state stored at the selected checkpoint---that is, the state from which the next attempt will resume.

From this context, it extracts a concise summary of the abandoned attempt, the
conditions that caused its failure, the hypotheses invalidated by the attempt, and the
subgoals or milestones it successfully reached. These outputs are used to update the
Attempt History and Failure Analysis fields of the Reflection Memory. The Task Objective
remains fixed, while the Environment Model is not modified by the reflection patcher:
environment knowledge is committed separately by the reviewer and therefore passes
through this update unchanged. Figure~\ref{fig:reflection} reproduces the role and rule
sections of the LLM-based reflection patcher prompt.

\section{Proofs for the Recovery-Space Analysis}
\label{app:theory}

This section provides the definitions and proofs underlying
Proposition~1 and Corollary~1 in Section~\ref{sec:theory}.
The comparison holds memory-update options fixed at each
rollback depth and varies only the admissible depths.

\paragraph{Recovery configurations.}
At a recovery event on branch step $t$, let $\mathcal{I}_t$
be the admissible checkpoint indices and define
\begin{equation}
\mathcal{K}_t
=
\{\,t-r \mid r\in\mathcal{I}_t\,\}.
\end{equation}
For $r=t-k$, let
$\mathbb{M}(\mathcal{M},\tau_{r:t})$ be the admissible
memory configurations associated with that restore point.
This family includes the unchanged memory $\mathcal{M}$
and the full update $\phi(\mathcal{M},\tau_{r:t})$
from Equation~\ref{eq:update}.
The proof does not require arbitrary combinations of
partially updated fields; any additional configurations
must be admitted equally by both policy classes.

The recovery operator is
\begin{equation}
\mathcal{R}_{k,\mathcal{M}^{+}}:
(s_t,h_t,\mathcal{M})
\longmapsto
(s_{t-k},h_{t-k},\mathcal{M}^{+}),
\qquad
(k,\mathcal{M}^{+})\in\mathcal{C}_t,
\label{eq:app_unified_operator}
\end{equation}
where
\begin{equation}
\mathcal{C}_t
=
\left\{
(k,\mathcal{M}^{+})
\,\middle|\,
k\in\mathcal{K}_t,\;
\mathcal{M}^{+}\in
\mathbb{M}(\mathcal{M},\tau_{t-k:t})
\right\}.
\end{equation}
The corresponding operator family is
\begin{equation}
\mathfrak{R}_{\mathrm{RIR}}(t)
=
\left\{
\mathcal{R}_{k,\mathcal{M}^{+}}
\,\middle|\,
(k,\mathcal{M}^{+})\in\mathcal{C}_t
\right\}.
\label{eq:rir_recovery_space}
\end{equation}
The dependence of the memory family on $k$ is explicit:
different restore points induce different abandoned suffixes.

\paragraph{Depth-restricted mechanisms.}
A restricted mechanism $j$ permits depths in
$\mathcal{K}_{j,t}\subseteq\mathcal{K}_t$ and therefore
has configuration set
\begin{equation}
S_{j,t}
=
\left\{
(k,\mathcal{M}^{+})\in\mathcal{C}_t
\,\middle|\,
k\in\mathcal{K}_{j,t}
\right\}.
\end{equation}
For every permitted depth, its memory options are identical
to those of the unrestricted class.
Examples include restart with
$\mathcal{K}_{j,t}=\{t\}$, fixed-depth rollback with
$\mathcal{K}_{j,t}=\{k_0\}$, and bounded-depth rollback with
$\mathcal{K}_{j,t}
=\{k\in\mathcal{K}_t:k\leq b\}$,
whenever these sets are admissible and nonempty.
A restriction need not be strict at every event.

\paragraph{Policy classes and comparison conditions.}
Let $\Pi_{\mathrm{RIR}}$ denote the idealized class of
history-dependent recovery policies with access to
$\mathcal{C}_t$, and let $\Pi_j$ denote the corresponding
class restricted to $S_{j,t}$.
Policies may randomize, but may use only information
available at the current decision.
The two classes otherwise have identical policy freedom.

Both classes share the initial task distribution, base agent,
available information, review process, checkpoint-restoration
semantics, non-recovery updates, and termination rules.
They also share the same budgets and cost accounting:
reproducing a configuration at the same history incurs the
same cost and induces the same subsequent transition law.
The restriction applies at every recovery history, with
nonempty feasible configuration sets whenever a recovery
decision is required.

Although indexed by the branch-local step $t$, the admissible
sets may depend on the full recovery history and remaining
budget. Budget counters are cumulative and are not reset
when execution returns to an earlier checkpoint.
The analysis concerns admissible policy behavior under these
shared rules; it does not assign different costs merely
because one policy class offers more choices.

\paragraph{Proof of Proposition~1.}
Take any $\pi_j\in\Pi_j$.
At each recovery history, construct a policy
$\widetilde{\pi}$ that assigns the same probabilities as
$\pi_j$ to configurations in $S_{j,t}$ and zero probability
to configurations in $\mathcal{C}_t\setminus S_{j,t}$.
Since $S_{j,t}\subseteq\mathcal{C}_t$, these decisions are
admissible for the unrestricted class.

Starting from the shared initial distribution, the two
policies induce the same distribution over subsequent
histories. To see this, condition on any common history.
At a recovery event, their configuration distributions agree;
between recovery events, the base agent and other transition
rules agree. Their conditional next-history distributions
and incurred costs are therefore identical.
Induction over chronological execution events establishes
equality of their trajectory and budget-consumption laws.
Consequently, every budget-feasible policy in $\Pi_j$
is also available in $\Pi_{\mathrm{RIR}}$, giving
\begin{equation}
\Pi_j\subseteq\Pi_{\mathrm{RIR}}.
\label{eq:policy_inclusion}
\end{equation}
This proves Proposition~1.

\paragraph{Proof of Corollary~1.}
For verifiable tasks, let $R_g\in\{0,1\}$ indicate success
and define $J(\pi)=\Pr_{\pi}(R_g=1)$ under the shared
task distribution and execution rules.
For a nonempty policy class $\Pi$, let
$V^\star(\Pi)=\sup_{\pi\in\Pi}J(\pi)$.
By Proposition~1,
\begin{equation}
V^\star(\Pi_{\mathrm{RIR}})
=
\sup_{\pi\in\Pi_{\mathrm{RIR}}}J(\pi)
\geq
\sup_{\pi\in\Pi_j}J(\pi)
=
V^\star(\Pi_j).
\end{equation}
No maximizing policy need exist for this inequality to hold.
The same argument applies to expected return whenever
the objective is well-defined and bounded.
This proves Corollary~1.

\section{Additional Experimental Details}
\label{app:experimental_setup}

\subsection{Dataset and Metric Details}
\label{app:datasets_metrics}

\textbf{ALFWorld.}
ALFWorld~\citep{shridhar2020alfworld} is a text-based embodied environment covering
six task families: pick-and-place, pick-two, clean-and-place, heat-and-place,
cool-and-place, and examine-under-light. We evaluate on test dataset from the
official split. A task is considered successful
only when the environment returns its success termination signal.

For an evaluation set containing $N$ tasks, we report Success Rate (SR):
\begin{equation}
\mathrm{SR}
=
\frac{1}{N}
\sum_{i=1}^{N}
\mathbb{I}[z_i=1],
\end{equation}
where $z_i$ indicates whether task $i$ is successfully completed. Timeouts,
execution errors, and unfinished episodes are counted as failures.

\textbf{ScienceWorld.}
ScienceWorld~\citep{wang2022scienceworld} is a text-based scientific environment
containing 30 task types with multiple environment variations. To avoid allowing
task categories with many variations to dominate the aggregate results, we sample
at most 10 test variations from each task type, resulting in 271 evaluation tasks. ScienceWorld assigns each trajectory a cumulative score $q_i\in[0,100]$. A task is
considered fully successful only when $q_i=100$. In addition to SR, we report the
average normalized Dense Reward (DR):
\begin{equation}
\mathrm{DR}
=
\frac{1}{N}
\sum_{i=1}^{N}
\frac{\max(0,q_i)}{100}.
\end{equation}
DR complements binary success by capturing partial progress on trajectories that do
not fully complete the task.

\textbf{GAIA.}
GAIA~\citep{mialon2024gaia} evaluates general-purpose agents on open-domain tasks
involving web search, code execution, file processing, and multi-tool reasoning.
We evaluate on the dataset after filtering image- and audio-dependent instances
from the public validation split and report results for overall performance. 

For GAIA, an independent evaluator receives the task question $x_i$, reference answer
$y_i^\star$, and agent prediction $\hat{y}_i$, and returns a binary correctness label:
\begin{equation}
z_i
=
J_{\mathrm{eval}}
(x_i,y_i^\star,\hat{y}_i)
\in\{0,1\}.
\end{equation}
GAIA success rate is then computed over the complete evaluation subset using the same
SR definition above. The evaluator has no access to the agent trajectory and does not
participate in agent execution, review, restore localization, or Reflection Memory
updates. Empty answers, malformed outputs, timeouts, and unrecoverable execution errors
are counted as failures.

\subsection{LLM-as-a-Judge Protocol}
\label{app:judge_protocol}

For GAIA, final-answer correctness is evaluated by an independent
Qwen3.8-Max model. The evaluator is invoked only once after an episode terminates
and is completely separated from the acting agent and all recovery components.
Its input contains only the original
task, the reference answer, the predicted answer, and minimal metadata required to
interpret the expected answer format. The evaluator is instructed to ignore superficial differences in wording and judge
whether the prediction is semantically consistent with the reference answer while
satisfying all required answer constraints, as shown in Figure~\ref{fig:judge-prompt}.

\begin{figure}[t]
\centering
\begin{minipage}{\textwidth}
\begin{tcolorbox}[
  colback=white, colframe=black!55, boxrule=0.4pt, arc=1.5pt,
  left=3pt, right=3pt, top=2pt, bottom=2pt,
  fonttitle=\bfseries\footnotesize, coltitle=black, colbacktitle=black!8,
  title={GAIA Judge Prompt}
]
{\ttfamily\scriptsize
You are an impartial evaluator. Judge whether the prediction satisfies the task
given the reference answer. Accept equivalent wording, formatting, units, and
aliases where appropriate. Ignore superficial differences that do not affect
the correctness of the answer.

\medskip
Task: \{question\}\\
Reference answer: \{reference\_answer\}\\
Prediction: \{prediction\}\\
Metadata: \{task\_metadata\}

\medskip
Return a structured judgment containing:
\begin{itemize}
    \item \texttt{correct}: whether the prediction satisfies the task;
    \item \texttt{score}: confidence in the judgment, from 0 to 1;
    \item \texttt{reason}: a brief rationale for the judgment.
\end{itemize}
}
\end{tcolorbox}

\begin{tcolorbox}[
  colback=white, colframe=black!55, boxrule=0.4pt, arc=1.5pt,
  left=3pt, right=3pt, top=2pt, bottom=2pt,
  fonttitle=\bfseries\footnotesize, coltitle=black, colbacktitle=black!8,
  title={Judge Response Schema}
]
\begin{lstlisting}[style=rirjson]
{
  "type": "object",
  "required": ["correct", "score", "reason"],
  "properties": {
    "correct": {
      "type": "boolean",
      "description":
        "Whether the prediction satisfies the task."
    },
    "score": {
      "type": "number",
      "minimum": 0.0,
      "maximum": 1.0,
      "description":
        "Confidence in the correctness judgment."
    },
    "reason": {
      "type": "string",
      "description":
        "Brief evaluation rationale."
    }
  }
}
\end{lstlisting}
\end{tcolorbox}
\end{minipage}

\caption{Prompt used by the independent LLM-as-a-Judge for GAIA final-answer
evaluation. The judge receives the task, reference answer, prediction, and task
metadata, and does not observe the agent trajectory or participate in execution
or recovery.}
\label{fig:judge-prompt}
\end{figure}
To prevent evaluation leakage, reference answers are never exposed to the acting
agent, reviewer, selector, reflection memory patcher, or any baseline controller. Judge calls
are used exclusively for evaluation and are therefore excluded from the inference cost
of the compared methods.

\subsection{Baseline Implementation Details}
\label{app:baseline_details}

All baselines use the same backbone model, environment interface, task prompt, and
tool set as RIR within each experimental condition. Differences are restricted to
the recovery, reflection, or search mechanism required by each method. Whenever
possible, we follow the official implementation and recommended configuration of
the corresponding method. Components such as critics or evaluators use the same
backbone as the acting agent unless otherwise stated.

\textbf{ReAct.}
ReAct~\citep{yao2022react} serves as the base agent without an explicit recovery
mechanism. The agent alternates between reasoning and environment actions using the
current task context, without additional reflection memory or state rollback.

\textbf{Self-Refine.}
Self-Refine~\citep{madaan2023self} performs information-level correction without
restoring the environment state. We periodically invoke a critic over the current
trajectory and inject its feedback into subsequent agent context. A refinement is
triggered every 6 steps, with at most 8 refinement rounds per task. Once the critic determines that no
correction is required, no additional refinement is performed until the next
scheduled review.

\textbf{Reflexion.}
Reflexion~\citep{shinn2023reflexion} is implemented using repeated trials with
cross-trial verbal reflection. Each task receives at most 2 trials. After a failed trial, the environment is reset to its
initial state and the model summarizes the failed trajectory into a textual reflection
that is provided to the next trial. Agent calls from all trials contribute to the
same cumulative interaction budget.

\textbf{LATS.}
LATS~\citep{zhou2023language} explores and evaluates multiple candidate trajectories
through tree search. We use an expansion width of 3, at most
8 search iterations, and a simulation depth of 5. All model calls generated by branch expansion, evaluation, and
simulation are included in the cumulative interaction cost.

\textbf{GA-Rollback.}
GA-Rollback~\citep{li2025generator} periodically inspects the generated trajectory and
restores execution to an earlier state when an unreliable step is detected. Following
the published protocol, we use
$\mathrm{wait}\text{-}k=6$ and a maximum rollback span of
4 per task.

\textbf{RIR.}
The full RIR configuration enables Hybrid Adaptive Review, Coarse-to-Fine Restore
Localization, and Rollback-Consistent Reflection Memory. For each backbone condition,
the acting agent, reviewer, selector, and reflection memory patcher all use the same LLM.
The initial review horizon is 6, with an adaptive range of
$\texttt{$[4,8]$}$.

\subsection{Implementation and Fairness Details}
\label{app:implementation_details}

\textbf{Model configuration.}
We evaluate each method using Qwen3-14B, DeepSeek-V3, and Qwen3.8-27B as the backbone models.
Within a given backbone condition, the acting agent and all method-specific auxiliary
components use the same model. In particular, RIR does not rely on a stronger model
for review, restore localization, or reflection extraction. GAIA final-answer
evaluation uses the independent Qwen3.8-Max judge described in
Appendix~\ref{app:judge_protocol}.

\begin{table}[t]
\centering
\small
\caption{
Model assignments used in our experiments. Recovery components always use the same
backbone as the acting agent; the independent judge is used only for final GAIA
evaluation.
}
\label{tab:model_assignments}
\begin{tabular}{lccc}
\toprule
Benchmark & Agent Backbone & Recovery Components & Judge \\
\midrule

ALFWorld
& Qwen3-14B / DeepSeek-V3 / Qwen3.8-27B
& Same as backbone
& Environment \\

ScienceWorld
& Qwen3-14B / DeepSeek-V3 / Qwen3.8-27B
& Same as backbone
& Environment \\

GAIA
& Qwen3-14B / DeepSeek-V3 / Qwen3.8-27B
& Same as backbone
& Qwen3.8-Max \\

\bottomrule
\end{tabular}
\end{table}

\textbf{State restoration.}
For ALFWorld and ScienceWorld, checkpoints are restored by resetting the environment
and replaying the retained action prefix when a native persistent snapshot is not
available. 
GAIA requires additional handling because tool calls may create or modify local files.
A checkpoint therefore includes both the agent context and the task workspace state.
When restoring a checkpoint, files created exclusively by the abandoned suffix are
removed and modified files are restored to their checkpointed versions, preventing
filesystem side effects from leaking across branches. Restoration is limited to controlled execution state, including the environment state, agent context, and local task workspace supported by the restoration interface. It does not imply reversibility of arbitrary external actions: restoring a checkpoint cannot, by itself, retract an email already sent or undo another irreversible action committed to an external system.

\textbf{GAIA tools.}
All methods receive the same set of web search, web browsing, code execution, and file
I/O tools. Tool outputs, attachment paths, and attachment-size limits are held fixed
across methods. Neither prompts nor tool interfaces expose the reference answer during
agent execution.

\textbf{Execution budgets.}
We use benchmark-specific interaction limits to account for differences in task
horizon and environment complexity. On ALFWorld, we set the maximum number of
interaction steps to 50 and allow at most 20 rollbacks per task. On ScienceWorld,
the maximum number of interaction steps is 100, with the rollback budget likewise
set to 20. On GAIA, we use a maximum of 24 interaction steps and allow at most
20 rollbacks per task. Once either the interaction-step budget or the rollback
budget is exhausted, no further interaction or rollback is permitted. All reported experimental results are averaged over three independent runs.

\section{Additional Results}
\label{d}
\subsection{Complete Efficiency Results}
\label{app:efficiency_details}

\begin{figure*}[t]
    \centering

    \begin{minipage}[t]{0.485\textwidth}
        \centering
        \includegraphics[width=\linewidth]
        {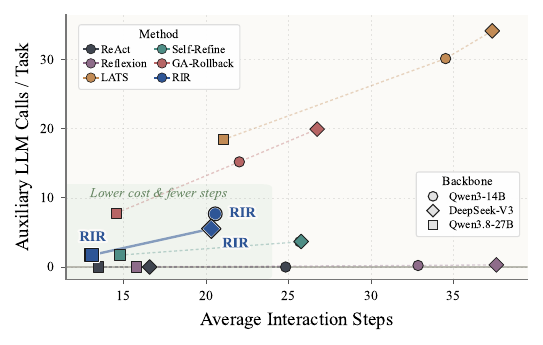}

        \vspace{-1mm}
        {\small (a) ALFWorld}
    \end{minipage}
    \hfill
    \begin{minipage}[t]{0.485\textwidth}
        \centering
        \includegraphics[width=\linewidth]
        {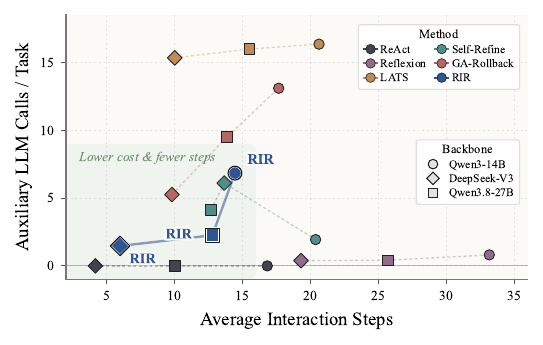}

        \vspace{-1mm}
        {\small (b) GAIA}
    \end{minipage}

    \caption{
    Additional efficiency comparisons on ALFWorld and GAIA across multiple backbones. The horizontal axis denotes average interaction steps and
    the vertical axis denotes auxiliary LLM calls per task. Colors represent
    methods and marker shapes represent LLM backbones. Lower-left corresponds
    to lower interaction and auxiliary inference overhead. For Reflexion, interaction steps are summed across up to two trials. On GAIA, the 24-step limit applies separately to each trial, so the cumulative number of steps per task can exceed 24.
    }
    \label{fig:additional_efficiency}
\end{figure*}

Figure~\ref{fig:additional_efficiency} provides the corresponding efficiency
comparisons on ALFWorld and GAIA. As in the ScienceWorld results in
Figure~\ref{fig:scienceworld_efficiency}, the horizontal axis denotes the average
number of interaction steps and the vertical axis denotes auxiliary LLM calls per
task. Colors distinguish methods, while marker shapes distinguish LLM backbones.
Points closer to the lower-left indicate lower test-time overhead.

Table~\ref{tab:additional_rollback} reports the corresponding rollback frequency.
Despite allowing up to 20 rollbacks per task, RIR invokes recovery infrequently
in both benchmarks. It uses fewer rollbacks than GA-Rollback in five of the six
backbone--benchmark settings; the only exception is ALFWorld with DeepSeek-V3,
where both methods already exhibit low rollback frequency.

\begin{table}[t]
\centering
\caption{
Average number of state rollbacks per task on ALFWorld and GAIA.
}
\label{tab:additional_rollback}

\small
\setlength{\tabcolsep}{7pt}
\renewcommand{\arraystretch}{1.10}

\begin{tabular}{llcc}
\toprule

\textbf{Backbone}
& \textbf{Method}
& \textbf{ALFWorld}
& \textbf{GAIA}
\\

\midrule

\multirow{2}{*}{Qwen3-14B}
& GA-Rollback
& 0.490
& 1.290
\\

& \textbf{RIR}
& \textbf{0.410}
& \textbf{0.546}
\\

\midrule

\multirow{2}{*}{DeepSeek-V3}
& GA-Rollback
& \textbf{0.224}
& 0.787
\\

& \textbf{RIR}
& 0.657
& \textbf{0.184}
\\

\midrule

\multirow{2}{*}{Qwen3.8-27B}
& GA-Rollback
& 0.049
& 0.880
\\

& \textbf{RIR}
& \textbf{0.036}
& \textbf{0.408}
\\

\bottomrule
\end{tabular}
\end{table}

\subsection{Additional Ablation Results}

\begin{table}[t]
\centering
\caption{
Ablation results on ScienceWorld with Qwen3-14B.
\textbf{SR} denotes success rate and \textbf{DR} denotes normalized dense reward.
\textbf{Avg.\ Steps}, \textbf{Rev.}, \textbf{RB}, and \textbf{Aux.} report per-task averages of environment interaction steps,
reviewer invocations, executed rollbacks, and auxiliary LLM calls, respectively.
Executed rollbacks restore both the environment state and branch-local trajectory to an earlier
checkpoint.
}
\label{tab:qwen_ablation_appendix}
\small
\setlength{\tabcolsep}{12pt}
\begin{tabular}{@{}lcccccc@{}}
\toprule
\textbf{Variant}
& \textbf{SR$\uparrow$}
& \textbf{DR$\uparrow$}
& \textbf{Avg.\ Steps}
& \textbf{Rev.}
& \textbf{RB}
& \textbf{Aux.} \\
\midrule

\rowcolor{black!4}
\multicolumn{7}{c}{\textit{Review-trigger variants}} \\

Agent-only
& 34.32 & 0.435 & 53.4 & 5.39 & 0.69 & 7.70 \\

Sched.-only (fix.)
& 33.58 & 0.491 & 45.8 & 7.10 & 0.54 & 9.01 \\

Sched.-only (adap.)
& 34.32 & 0.468 & 46.9 & 8.48 & 0.75 & 11.27 \\

Hybrid (fixed)
& 36.90 & 0.445 & 50.2 & 8.20 & 0.80 & 10.96 \\

\addlinespace[2pt]

\rowcolor{black!4}
\multicolumn{7}{c}{\textit{Restore-localization variants}} \\

No rollback
& 33.95 & 0.441 & 43.1 & 8.30 & / & 11.44 \\

Restart
& 35.79 & 0.454 & 54.4 & 10.56 & 1.17 & 14.07 \\

Direct selection
& 36.16 & 0.490 & 49.4 & 9.27 & 0.86 & 12.80 \\

\addlinespace[2pt]

\rowcolor{black!4}
\multicolumn{7}{c}{\textit{Recovery-component variants}} \\

Reflection only
& 33.95 & 0.441 & 43.1 & 8.30 & / & 11.44 \\

Rollback only
& 35.42 & 0.439 & 48.7 & 9.47 & 1.04 & 13.21 \\

\midrule
\rowcolor{rirhi}
\textbf{RIR}
& 36.67 & 0.448 & 51.5 & 9.90 & 0.98 & 13.33 \\

\bottomrule
\end{tabular}
\end{table}

\begin{table}[t]
\centering
\caption{
Ablation results on ScienceWorld with Qwen3.8-27B.
\textbf{SR} denotes success rate and \textbf{DR} denotes normalized dense reward.
\textbf{Avg.\ Steps}, \textbf{Rev.}, \textbf{RB}, and \textbf{Aux.} report per-task averages of environment interaction steps,
reviewer invocations, executed rollbacks, and auxiliary LLM calls, respectively.
Executed rollbacks restore both the environment state and branch-local trajectory to an earlier
checkpoint.
}
\label{tab:qwen27b_ablation_appendix}
\small
\setlength{\tabcolsep}{12pt}
\begin{tabular}{@{}lcccccc@{}}
\toprule
\textbf{Variant}
& \textbf{SR$\uparrow$}
& \textbf{DR$\uparrow$}
& \textbf{Avg.\ Steps}
& \textbf{Rev.}
& \textbf{RB}
& \textbf{Aux.} \\
\midrule

\rowcolor{black!4}
\multicolumn{7}{c}{\textit{Review-trigger variants}} \\

Agent-only
& 73.43 & 0.837 & 37.1 & 3.31 & 0.73 & 5.46 \\

Sched.-only (fix.)
& 73.80 & 0.844 & 36.2 & 5.65 & 0.97 & 8.59 \\

Sched.-only (adap.)
& 70.11 & 0.810 & 46.0 & 9.92 & 1.61 & 12.33 \\

Hybrid (fixed)
& 76.38 & 0.862 & 38.8 & 7.08 & 1.15 & 8.63 \\

\addlinespace[2pt]

\rowcolor{black!4}
\multicolumn{7}{c}{\textit{Restore-localization variants}} \\

No rollback
& 75.64 & 0.860 & 31.7 & 7.61 & / & 9.74 \\

Restart
& 71.59 & 0.831 & 49.1 & 8.15 & 1.54 & 12.47 \\

Direct selection
& 71.96 & 0.829 & 38.6 & 8.03 & 1.26 & 9.80 \\

\addlinespace[2pt]

\rowcolor{black!4}
\multicolumn{7}{c}{\textit{Recovery-component variants}} \\

Reflection only
& 75.64 & 0.860 & 31.7 & 7.61 & / & 9.74 \\

Rollback only
& 74.17 & 0.852 & 41.8 & 7.28 & 1.89 & 11.51 \\

\midrule
\rowcolor{rirhi}
\textbf{RIR}
& 79.70 & 0.872 & 39.2 & 7.11 & 1.27 & 10.70 \\

\bottomrule
\end{tabular}
\end{table}

Across the three backbones, RIR maintains a consistent advantage over the non-RIR
baselines in the main results (Table~\ref{tab:main_results}). To examine whether this
advantage also holds at the level of individual design choices, we report
component-level ablations on ScienceWorld for Qwen3-14B
(Table~\ref{tab:qwen_ablation_appendix}) and Qwen3.8-27B
(Table~\ref{tab:qwen27b_ablation_appendix}), complementing the DeepSeek-V3 ablations
in the main text. The picture is consistent across all three backbones: the default
RIR configuration---hybrid adaptive review, coarse-to-fine localization, and
rollback-consistent reflection---attains the best or a closely comparable SR in every
ablation group, and its margin over the constrained variants widens as the backbone
becomes more capable. Because RIR delegates the open-ended \emph{when}, \emph{where},
and \emph{what} decisions to the reviewer, a stronger reviewer can better exploit the
additional flexibility, whereas on the weaker Qwen3-14B the more autonomous and the
more constrained variants converge to comparable performance.

\textbf{When to review.}
Combining agent-initiated and scheduled review is beneficial on both backbones: the
two hybrid variants outperform every single-trigger variant. On Qwen3.8-27B, the full
adaptive RIR is clearly best (79.70\% SR, 0.872 DR), exceeding Hybrid~(fixed) by 3.32
points in SR. On Qwen3-14B, the two hybrid variants are essentially tied---Hybrid~(fixed)
reaches 36.90\% SR and RIR 36.67\%---while RIR retains a marginally higher DR
(0.448 vs.\ 0.445). Adaptive triggering asks the reviewer to judge whether the
trajectory has left a recoverable path and whether immediate intervention is preferable
to continued exploration; a stronger reviewer answers these questions reliably enough
to turn the added flexibility into a clear gain, whereas on Qwen3-14B adaptive and fixed
scheduling perform comparably.

\textbf{Where to recover.}
Coarse-to-fine localization yields the best SR on both backbones. On Qwen3.8-27B, RIR
reaches 79.70\% SR and 0.872 DR, well above direct selection (71.96\%) and no rollback
(75.64\%). On Qwen3-14B, RIR again attains the highest SR (36.67\%), ahead of direct
selection (36.16\%), restart (35.79\%), and no rollback (33.95\%); the last confirms
that reflection without state restoration cannot reliably recover failed trajectories.
Direct selection obtains a slightly higher DR on Qwen3-14B (0.490 vs.\ 0.448) but at a
lower SR, so the two-stage selector provides the better success--progress balance. Its
advantage is larger on the stronger backbone, where region localization and checkpoint
selection are both more reliable and therefore compound rather than conflict.

\textbf{Rollback vs.\ Reflection.}
Coupling state restoration with reflection is preferred on both backbones and on both
metrics. On Qwen3.8-27B, RIR reaches 79.70\% SR and 0.872 DR, above reflection only
(75.64\% SR, 0.860 DR) and rollback only (74.17\% SR, 0.852 DR). On Qwen3-14B, RIR
leads with 36.67\% SR and 0.448 DR, exceeding rollback only (35.42\% SR, 0.439 DR) and
reflection only (33.95\% SR, 0.441 DR). Restoring the environment and trajectory removes
the consequences of an earlier erroneous action, while rollback-consistent reflection
preserves transferable knowledge across attempts; the two mechanisms are complementary
rather than redundant, and their combination is the strongest recovery-component
configuration regardless of backbone capability.

Taken together, these ablations reinforce rather than qualify the design of RIR. On all
three backbones, the default configuration is the best or a closely comparable choice in
every ablation group, and no constrained variant dominates it: where a single-axis
variant achieves a marginally higher DR on Qwen3-14B, it does so at a clear cost in SR.
The advantage grows with backbone capability, consistent with the view that adaptive
review, coarse-to-fine localization, and rollback-consistent reflection give a more
capable reviewer more useful room to act. RIR is therefore best viewed as a general
recovery framework whose full configuration transfers across backbone scales, with the
largest gains realized on stronger models.

\subsection{Comparison with Non-Rollback Planning and Reflection Methods}
\label{app:nonrollback_comparison}

We extend the evaluation with HiPlan~\citep{li2025hiplan} and ReflAct~\citep{kim2025reflact},
two methods that improve forward execution without restoring earlier execution
states. HiPlan uses hierarchical global--local planning, whereas ReflAct uses
goal-state reflection to guide subsequent action selection.
We implement both methods within our evaluation framework and compare them
with RIR on ALFWorld, ScienceWorld, and GAIA using Qwen3.8-27B and DeepSeek-V3.
Within each backbone setting, the methods use the same task environments,
task instances, and available task tools.
Table~\ref{tab:nonrollback_comparison} reports success rates and their
unweighted average across the three benchmarks. RIR achieves the highest success rate in all six backbone--benchmark settings.
Relative to the stronger added baseline by average success rate, ReflAct,
RIR improves average performance by 2.81 percentage points on Qwen3.8-27B
and 5.98 points on DeepSeek-V3.
On Qwen3.8-27B, RIR exceeds the better of the two added baselines by
2.24 points on ALFWorld, 4.05 points on ScienceWorld, and 0.65 points on GAIA.
The corresponding improvements on DeepSeek-V3 are
5.97, 7.88, and 1.98 points.

\setlength{\tabcolsep}{12pt}
\begin{table}[t]
    \centering

    \caption{
        Comparison with non-rollback planning and reflection methods.
        All results are success rates (\%).
        Avg.\ is the unweighted mean across the three benchmarks.
        Bold denotes the best result within each backbone.
    }
    \label{tab:nonrollback_comparison}
        \small
        \begin{tabular}{llcccr}
            \hline
            Backbone & Method & ALFWorld & ScienceWorld & GAIA & Avg. \\
            \hline
            Qwen3.8-27B
                & HiPlan
                & 97.01 & 73.80 & 49.34 & 73.38 \\
                & ReflAct
                & 95.52 & 75.65 & 50.01 & 73.73 \\
                & RIR (Ours)
                & \textbf{99.25}
                & \textbf{79.70}
                & \textbf{50.66}
                & \textbf{76.54} \\
            \hline
            DeepSeek-V3
                & HiPlan
                & 83.58 & 68.87 & 33.57 & 62.01 \\
                & ReflAct
                & 88.06 & 66.79 & 35.52 & 63.45 \\
                & RIR (Ours)
                & \textbf{94.03}
                & \textbf{76.75}
                & \textbf{37.50}
                & \textbf{69.43} \\
            \hline
        \end{tabular}
    
\end{table}

\subsection{Direct Comparison with GA-Rollback + Reflexion}
\label{app:ga_reflexion_comparison}

We additionally compare RIR with GA-Rollback + Reflexion, a closely related
combination of rollback and verbal reflection.
Our framework implementation combines GA-Rollback-style recovery within a trial
with Reflexion-style memory across trials: after an unsuccessful trial,
the task restarts with a reflection summarizing the preceding attempt.
We evaluate this combination on ALFWorld, ScienceWorld, and GAIA using
Qwen3.8-27B and DeepSeek-V3. This experiment complements the internal \emph{Rollback only} ablation.
Rollback only retains RIR's recovery controller while withholding the reflection
briefing from the acting agent; it therefore examines the contribution of
generator-facing reflection within RIR.
The present comparison instead evaluates RIR against a separate method that
already includes both rollback and reflection.
Table~\ref{tab:ga_reflexion_comparison} reports the results. RIR improves the average success rate by 1.75 percentage points on Qwen3.8-27B
and 8.21 points on DeepSeek-V3.
On DeepSeek-V3, RIR improves all three benchmarks, with gains of
14.18 points on ALFWorld, 8.48 points on ScienceWorld, and 1.97 points on GAIA.
On Qwen3.8-27B, GA-Rollback + Reflexion leads by 0.66 points on GAIA,
while RIR leads by 1.49 points on ALFWorld and 4.42 points on ScienceWorld,
retaining the higher overall average.

\setlength{\tabcolsep}{12pt}
\begin{table}[t]
    \centering
    \caption{
        Direct comparison with GA-Rollback + Reflexion.
        All results are success rates (\%).
        Avg.\ is the unweighted mean across the three benchmarks.
        Bold denotes the best result within each backbone.
    }
    \label{tab:ga_reflexion_comparison}
    \small
        \begin{tabular}{llcccr}
            \hline
            Backbone & Method & ALFWorld & ScienceWorld & GAIA & Avg. \\
            \hline
            Qwen3.8-27B
                & GA-Rollback + Ref.
                & 97.76
                & 75.28
                & \textbf{51.32}
                & 74.79 \\
                & RIR (Ours)
                & \textbf{99.25}
                & \textbf{79.70}
                & 50.66
                & \textbf{76.54} \\
            \hline
            DeepSeek-V3
                & GA-Rollback + Ref.
                & 79.85 & 68.27 & 35.53 & 61.22 \\
                & RIR (Ours)
                & \textbf{94.03}
                & \textbf{76.75}
                & \textbf{37.50}
                & \textbf{69.43} \\
            \hline
        \end{tabular}
\end{table}

\section{Case Studies}
\label{app:case_studies}

We complement the aggregate results with a qualitative analysis of five representative
trajectories, illustrating both successful recovery and characteristic failure modes of RIR.
The selected cases span ALFWorld, ScienceWorld, and GAIA and are generated with
DeepSeek-V3. We include three successful trajectories in which rollback contributes to
subsequent task completion (Section~\ref{sec:case_good}) and two unsuccessful trajectories
that expose limitations in restore localization and reflection
(Section~\ref{sec:case_bad}).

For successful cases, we consider trajectories that complete the task with at least one
rollback, since trajectories without recovery provide limited evidence about the rollback
mechanism itself. For failure analysis, we focus on trajectories with repeated recovery
attempts, which make the source of unsuccessful recovery easier to isolate. To expose the
full recovery dynamics of these trajectories, we use an enlarged maximum-step budget for
the qualitative analysis, allowing repeated rollback-and-retry cycles to unfold more
completely; this diagnostic setting is used only for the case studies and does not affect
the main experimental results. Quoted trajectory content, including selector reasoning,
reflection outputs, and Reflection Memory entries, is reproduced directly from the
corresponding execution logs. Each case is followed by an analysis of the recovery behavior
it illustrates, and Section~\ref{sec:case_summary} summarizes the common patterns across all
five trajectories.

Trajectory indices are branch-local and therefore may repeat after rollback; each rollback
transition explicitly indicates the checkpoint from which execution resumes.


\subsection{Good Cases}
\label{sec:case_good}


\subsubsection{G1 --- ALFWorld, committed state error}

\begin{casestudy}{G1 \quad ALFWorld}

{\small\textbf{Task.} \texttt{put a soapbottle in toilet}}

\begin{casephase}{CaseNeutral}{CaseNeutralBG}{Initial exploration}{idx 0--5}
\begin{tabular}{@{}r@{\hspace{0.6em}}l@{\hspace{0.5em}}p{\dimexpr\linewidth-4.2em\relax}@{}}
{\scriptsize\color{black!55}0} & {\scriptsize\bfseries ACT} & \texttt{go to cabinet 1}\\
                               & {\scriptsize\bfseries OBS} & You arrive at cabinet 1. The cabinet 1 is closed.\\[3pt]
{\scriptsize\color{black!55}1} & {\scriptsize\bfseries ACT} & \texttt{open cabinet 1}\\
                               & {\scriptsize\bfseries OBS} & You open the cabinet 1. The cabinet 1 is open. In it, you see a candle 2, and a cloth 2.\\[3pt]
{\scriptsize\color{black!55}2} & {\scriptsize\bfseries ACT} & \texttt{go to cabinet 2}\\
                               & {\scriptsize\bfseries OBS} & You arrive at cabinet 2. The cabinet 2 is closed.\\[3pt]
{\scriptsize\color{black!55}3} & {\scriptsize\bfseries ACT} & \texttt{open cabinet 2}\\
                               & {\scriptsize\bfseries OBS} & You open the cabinet 2. The cabinet 2 is open. In it, you see a candle 3, and a candle 1.\\[3pt]
{\scriptsize\color{black!55}4} & {\scriptsize\bfseries ACT} & \texttt{go to cabinet 3}\\
                               & {\scriptsize\bfseries OBS} & You arrive at cabinet 3. The cabinet 3 is closed.\\[3pt]
{\scriptsize\color{black!55}5} & {\scriptsize\bfseries ACT} & \texttt{open cabinet 3}\\
                               & {\scriptsize\bfseries OBS} & You open the cabinet 3. The cabinet 3 is open. In it, you see a spraybottle 1.
\end{tabular}
\end{casephase}

\begin{casephase}{CaseFail}{CaseFailBG}{Abandoned branch}{idx 6--9}
\begin{tabular}{@{}r@{\hspace{0.6em}}l@{\hspace{0.5em}}p{\dimexpr\linewidth-4.2em\relax}@{}}
{\scriptsize\color{black!55}6} & {\scriptsize\bfseries ACT} & \texttt{take spraybottle 1 from cabinet 3}\\
                               & {\scriptsize\bfseries OBS} & You pick up the spraybottle 1 from the cabinet 3.\\[3pt]
{\scriptsize\color{black!55}7} & {\scriptsize\bfseries ACT} & \texttt{go to toilet 1}\\
                               & {\scriptsize\bfseries OBS} & You arrive at toilet 1. On the toilet 1, you see a soapbar 2, and a toiletpaper 2.\\[3pt]
{\scriptsize\color{black!55}8} & {\scriptsize\bfseries ACT} & \texttt{move spraybottle 1 to toilet 1}\\
                               & {\scriptsize\bfseries OBS} & \textit{``You move the spraybottle 1 to the toilet 1.''}\\[3pt]
{\scriptsize\color{black!55}9} & {\scriptsize\bfseries ACT} & {[}ROLLBACK requested{]}\\
                               & {\scriptsize\bfseries WHY} & \textit{``The task was to put a soapbottle in the toilet, but the spraybottle was moved instead. The soapbottle was not found in any of the cabinets checked so far.''}
\end{tabular}
\end{casephase}

{\scriptsize\color{CaseFail}
$\blacktriangleright$
The action at idx 8 has already taken effect in the environment---this is not a rejected
or no-op action.
}

\begin{casephase}{CaseRollback}{CaseRollbackBG}{Recovery approved}{$C_9 \to C_5$}
\textit{``The root cause of the failure was the decision to take the spraybottle from cabinet 3 at step 6.
Restoring to step 5 (before opening cabinet 3) allows for a re-evaluation of the contents of cabinet 3
and avoids the mistake of taking the wrong item.
This checkpoint is the earliest and latest that avoids the error, and it has no prior rollbacks.''}
\end{casephase}

\begin{casephase}{CaseMemory}{CaseMemoryBG}{Reflection Memory retained across rollback}{4 entries}
{\scriptsize\bfseries OBSERVED}\;
cabinet 1: candle 2, cloth 2
\quad\textbar\quad
cabinet 2: candle 3, candle 1
\quad\textbar\quad
cabinet 3: spraybottle 1\\[2pt]
{\scriptsize\bfseries AFFORDANCE}\;
opening a cabinet: requires the cabinet to be closed
\end{casephase}

\begin{casephase}{CaseRecover}{CaseRecoverBG}{Recovered branch}{idx 5--9}
\begin{tabular}{@{}r@{\hspace{0.6em}}l@{\hspace{0.5em}}p{\dimexpr\linewidth-4.2em\relax}@{}}
{\scriptsize\color{black!55}5} & {\scriptsize\bfseries ACT} & \texttt{open cabinet 3}
                               \hfill{\scriptsize\color{black!45}(replayed)}\\
                               & {\scriptsize\bfseries OBS} & You open the cabinet 3. The cabinet 3 is open. In it, you see a spraybottle 1.\\[3pt]
{\scriptsize\color{black!55}6} & {\scriptsize\bfseries ACT} & \texttt{go to countertop 1}
                               \hfill{\scriptsize\color{CaseRecover}(unexplored)}\\
                               & {\scriptsize\bfseries OBS} & You arrive at countertop 1. On the countertop 1, you see a soapbottle 1, and a spraybottle 2.\\[3pt]
{\scriptsize\color{black!55}7} & {\scriptsize\bfseries ACT} & \texttt{take soapbottle 1 from countertop 1}\\
                               & {\scriptsize\bfseries OBS} & You pick up the soapbottle 1 from the countertop 1.\\[3pt]
{\scriptsize\color{black!55}8} & {\scriptsize\bfseries ACT} & \texttt{go to toilet 1}\\
                               & {\scriptsize\bfseries OBS} & You arrive at toilet 1. On the toilet 1, you see a soapbar 2, and a toiletpaper 2.\\[3pt]
{\scriptsize\color{black!55}9} & {\scriptsize\bfseries ACT} & {[}FINAL{]} \texttt{move soapbottle 1 to toilet 1}\\
                               & {\scriptsize\bfseries OBS} & You move the soapbottle 1 to the toilet 1.
\end{tabular}
\end{casephase}

{\scriptsize\color{CaseRecover}
$\blacktriangleright$
Previously searched cabinets are not revisited.
}
\hfill
\textcolor{CaseSuccess}{\textbf{\checkmark\ success}}

\end{casestudy}

\captionof{figure}{
\textbf{G1.}
RIR recovers from a committed state error in ALFWorld.
Rollback restores the environment to the state before the incorrect object was taken,
while previously acquired exploration knowledge remains available through Reflection Memory.
The recovered branch then searches an unexplored location and completes the task.
}
\label{fig:case_g1}

\textbf{Analysis.}
G1 provides a direct example of RIR's central asymmetry between execution state and
persistent knowledge. The incorrect action at idx 8 has already modified the environment,
so textual correction alone cannot undo its consequence. Rolling back to $C_5$ removes
the committed spraybottle branch, while Reflection Memory preserves what was learned from
the preceding exploration. After replaying the checkpointed cabinet action, the agent does
not return to cabinets 1 or 2 and instead explores the previously unvisited countertop,
where it finds the correct soapbottle. Recovery therefore avoids both failure modes of the
two simpler alternatives: continuing from the corrupted state and restarting without
retaining useful information from the failed attempt.


\subsubsection{G2 --- ScienceWorld, precondition restoration and self-correction}

\begin{casestudy}{G2 \quad ScienceWorld}

{\small\textbf{Task.}
\texttt{Find a(n) non-living thing. First, focus on the thing. Then, move it to the orange box in the bedroom.}
}

\begin{casephase}{CaseNeutral}{CaseNeutralBG}{Initial exploration}{idx 0--5}
\begin{tabular}{@{}r@{\hspace{0.6em}}l@{\hspace{0.5em}}p{\dimexpr\linewidth-4.2em\relax}@{}}
{\scriptsize\color{black!55}0} & {\scriptsize\bfseries ACT} & \texttt{teleport to bedroom}\\
                               & {\scriptsize\bfseries OBS} & You teleport to the bedroom.\\[3pt]
{\scriptsize\color{black!55}1} & {\scriptsize\bfseries ACT} & \texttt{look around}\\
                               & {\scriptsize\bfseries OBS} & This room is called the bedroom. In it, you see: \ldots\ a finger painting\ a orange box (containing nothing)\ a table. On the table is: nothing.\\[3pt]
{\scriptsize\color{black!55}2} & {\scriptsize\bfseries ACT} & \texttt{look at orange box}\\
                               & {\scriptsize\bfseries OBS} & a orange box (containing nothing)\\[3pt]
{\scriptsize\color{black!55}3} & {\scriptsize\bfseries ACT} & \texttt{teleport to workshop}\\
                               & {\scriptsize\bfseries OBS} & You teleport to the workshop.\\[3pt]
{\scriptsize\color{black!55}4} & {\scriptsize\bfseries ACT} & \texttt{look at table}\\
                               & {\scriptsize\bfseries OBS} & a table. On the table is: a battery\ a blue wire\ a green wire\ a orange wire\ a red light bulb, which is off\ \ldots\\[3pt]
{\scriptsize\color{black!55}5} & {\scriptsize\bfseries ACT} & \texttt{focus on battery}\\
                               & {\scriptsize\bfseries OBS} & You focus on the battery.
\end{tabular}
\end{casephase}

\begin{casephase}{CaseFail}{CaseFailBG}{Misdirected branch}{idx 6--18}
\begin{tabular}{@{}r@{\hspace{0.6em}}l@{\hspace{0.5em}}p{\dimexpr\linewidth-4.2em\relax}@{}}
{\scriptsize\color{black!55}6} & {\scriptsize\bfseries ACT} & \texttt{move battery to orange box in bedroom}\\
                               & {\scriptsize\bfseries OBS} & \textit{``No known action matches that input.''}\\[3pt]
{\scriptsize\color{black!55}7} & {\scriptsize\bfseries ACT} & \texttt{teleport to bedroom}\\
                               & {\scriptsize\bfseries OBS} & You teleport to the bedroom.\\[3pt]
{\scriptsize\color{black!55}8} & {\scriptsize\bfseries ACT} & \texttt{move battery to orange box}\\
                               & {\scriptsize\bfseries OBS} & \textit{``No known action matches that input.''}\\[3pt]
{\scriptsize\color{black!55}9} & {\scriptsize\bfseries ACT} & {[}ROLLBACK requested{]} \casebadge{CaseBad}{denied}\\
                               & {\scriptsize\bfseries WHY} & \textit{``The command `move battery to orange box' is not recognized \ldots\ Need to find the correct syntax or object to move.''}\\[3pt]
{\scriptsize\color{black!55}10} & {\scriptsize\bfseries ACT} & \texttt{pick up battery}\\
                                & {\scriptsize\bfseries OBS} & \textit{``No known action matches that input.''}\\[3pt]
{\scriptsize\color{black!55}12} & {\scriptsize\bfseries ACT} & \texttt{look at table}\\
                                & {\scriptsize\bfseries OBS} & a table. On the table is: nothing
                                \hfill{\scriptsize\color{CaseFail}(bedroom table---battery is in the workshop)}\\[3pt]
\multicolumn{3}{@{}l@{}}{\scriptsize\color{black!45}\textit{idx 11, 13 omitted: \texttt{look at battery} (unrecognized), repeated \texttt{look around}}}\\[3pt]
{\scriptsize\color{black!55}14} & {\scriptsize\bfseries ACT} & \texttt{focus on finger painting}\\
                                & {\scriptsize\bfseries OBS} & You focus on the finger painting.\\[3pt]
{\scriptsize\color{black!55}15} & {\scriptsize\bfseries ACT} & \texttt{move finger painting to orange box}\\
                                & {\scriptsize\bfseries OBS} & \textit{``You move the finger painting to the orange box.''}
                                \hfill{\scriptsize\color{CaseFail}(succeeds environmentally)}\\[3pt]
\multicolumn{3}{@{}l@{}}{\scriptsize\color{black!45}\textit{idx 16--17: \texttt{think} $\times 2$ $\rightarrow$ ``No env action was executed.''}}\\[3pt]
{\scriptsize\color{black!55}18} & {\scriptsize\bfseries ACT} & {[}ROLLBACK requested{]}\\
                                & {\scriptsize\bfseries WHY} & \textit{``\ldots\ the task requires focusing on a non-living thing. The finger painting may not qualify, and the agent needs to revisit the workshop to find a valid non-living object.''}
\end{tabular}
\end{casephase}

{\scriptsize\color{CaseFail}
$\blacktriangleright$
Two independent failures are entangled here:
an unmet spatial precondition (the battery is not co-located with the agent),
and a self-identified semantic misjudgment
(finger painting may not qualify as ``non-living'').
Note that the environment reports the former as
\textit{``No known action matches that input''}---a \emph{syntax} error message---
which is why both the agent's request at idx 9 and the retained memory
attribute the failure to syntax rather than to location.
}

\begin{casephase}{CaseRollback}{CaseRollbackBG}{Recovery approved}{$C_{18} \to C_5$}
\textit{``The root cause of the failure is the incorrect focus on the finger painting,
which may not qualify as a non-living thing.
Restoring to step 5 allows the agent to correctly identify the battery as a non-living thing \ldots''}
\end{casephase}

\begin{casephase}{CaseMemory}{CaseMemoryBG}{Reflection Memory retained across rollback}{4 entries}
\begin{tabular}{@{}l@{\hspace{0.5em}}p{\dimexpr\linewidth-6.0em\relax}@{}}
{\scriptsize\bfseries OBSERVED}   & battery: is a non-living thing\\[2pt]
{\scriptsize\bfseries AFFORDANCE} & focusing on an object: requires the object to be present and visible\\
                                   & moving an object: requires correct syntax or method to execute\\
                                   & moving an object: requires correct syntax or method to execute, as demonstrated by the failed attempts to move the battery
                                   \hfill{\scriptsize\color{black!45}(near-duplicate)}
\end{tabular}
\end{casephase}

\begin{casephase}{CaseRecover}{CaseRecoverBG}{Recovered branch}{idx 5--8}
\begin{tabular}{@{}r@{\hspace{0.6em}}l@{\hspace{0.5em}}p{\dimexpr\linewidth-4.2em\relax}@{}}
{\scriptsize\color{black!55}5} & {\scriptsize\bfseries ACT} & \texttt{focus on battery}
                               \hfill{\scriptsize\color{black!45}(replayed)}\\
                               & {\scriptsize\bfseries OBS} & You focus on the battery.\\[3pt]
{\scriptsize\color{black!55}6} & {\scriptsize\bfseries ACT} & \texttt{pick up battery}
                               \hfill{\scriptsize\color{CaseRecover}(same command as idx 10)}\\
                               & {\scriptsize\bfseries OBS} & \textit{``You move the battery to the inventory.''}\\[3pt]
{\scriptsize\color{black!55}7} & {\scriptsize\bfseries ACT} & \texttt{teleport to bedroom}\\
                               & {\scriptsize\bfseries OBS} & You teleport to the bedroom.\\[3pt]
{\scriptsize\color{black!55}8} & {\scriptsize\bfseries ACT} & {[}FINAL{]} \texttt{move battery to orange box}\\
                               & {\scriptsize\bfseries OBS} & You move the battery to the orange box.
\end{tabular}
\end{casephase}

{\scriptsize\color{CaseRecover}
$\blacktriangleright$
The identical \texttt{pick up battery} command failed at idx 10 and succeeded at idx 6
after the rollback---the only change is spatial co-location, not knowledge.
The rollback therefore repaired a precondition that neither the agent's request
nor the selector's rationale had explicitly identified.
}
\hfill
\textcolor{CaseSuccess}{\textbf{\checkmark\ success}}

\end{casestudy}

\captionof{figure}{
\textbf{G2.}
RIR restores an unmet spatial precondition in ScienceWorld.
The same \texttt{pick up battery} command that fails while the agent is in the bedroom
succeeds after rollback restores the agent to the workshop.
The trajectory also contains both a denied and an approved rollback request,
showing that agent-initiated recovery remains subject to reviewer adjudication.
}
\label{fig:case_g2}

\textbf{Analysis.}
G2 shows that state restoration can recover a latent execution precondition even when the
failure diagnosis is incomplete. At idx 10, \texttt{pick up battery} fails after the agent
has moved from the workshop to the bedroom; after rollback, the identical command succeeds
from the restored workshop state. The environment reports the original failure as
\textit{``No known action matches that input,''} causing both the agent and Reflection Memory
to interpret it primarily as a syntax problem. Nevertheless, restore localization returns
execution to a state in which the missing spatial precondition is satisfied. The case also
illustrates Hybrid Adaptive Review: the first agent-issued rollback request is rejected,
whereas the later request is approved after the branch develops into a clearer semantic
misjudgment. Thus, RIR can benefit from state-level recovery even when the textual diagnosis
does not fully identify the underlying execution condition.


\subsubsection{G3 --- GAIA, horizon-triggered abandonment of a dead technical path}

\begin{casestudy}{G3 \quad GAIA}

{\small\textbf{Task.}
\textit{Which of the text elements under CATEGORIES in the XML would contain the one food in the spreadsheet that does not appear a second time under a different name?}
}

{\scriptsize\color{black!55}
\texttt{\$ZIP} denotes the task's attachment archive; long paths are abbreviated.
}

\begin{casephase}{CaseFail}{CaseFailBG}{Exhausted technical path}{idx 0--5, all \texttt{ACT}}
\begin{tabular}{@{}r@{\hspace{0.6em}}l@{\hspace{0.5em}}p{\dimexpr\linewidth-4.2em\relax}@{}}
{\scriptsize\color{black!55}0} & {\scriptsize\bfseries ACT} & \texttt{unzip -l "\$ZIP"}\\
                               & {\scriptsize\bfseries OBS} & \texttt{food\_duplicates.xls} (28160 B), \texttt{CATEGORIES.xml} (23933 B)\\[3pt]
{\scriptsize\color{black!55}1} & {\scriptsize\bfseries ACT} & \texttt{unzip -p "\$ZIP" "food\_duplicates.xls" \textgreater{} /tmp/f.xls \&\& file /tmp/f.xls}\\
                               & {\scriptsize\bfseries OBS} & Composite Document File V2 Document, \ldots\ Creating Application: Microsoft Excel
                               \hfill{\scriptsize\color{CaseFail}(legacy \texttt{.xls})}\\[3pt]
{\scriptsize\color{black!55}2} & {\scriptsize\bfseries ACT} & \texttt{pip install xlsx2csv \&\& xlsx2csv /tmp/f.xls /tmp/f.csv}\\
                               & {\scriptsize\bfseries OBS} & \textit{``/bin/sh: pip: command not found''}\\[3pt]
{\scriptsize\color{black!55}3} & {\scriptsize\bfseries ACT} & \texttt{python3 -c "import pandas as pd; df = pd.read\_excel('/tmp/f.xls'); print(df.to\_string())"}\\
                               & {\scriptsize\bfseries OBS} & \textit{``ModuleNotFoundError: No module named `xlrd' \ldots\ Missing optional dependency `xlrd'.''}\\[3pt]
{\scriptsize\color{black!55}4} & {\scriptsize\bfseries ACT} & \texttt{file /tmp/f.xls}
                               \hfill{\scriptsize\color{CaseFail}(identical to idx 1)}\\
                               & {\scriptsize\bfseries OBS} & Composite Document File V2 Document, \ldots
                               \hfill{\scriptsize\color{CaseFail}(byte-identical output)}\\[3pt]
{\scriptsize\color{black!55}5} & {\scriptsize\bfseries ACT} & \texttt{ssconvert /tmp/f.xls /tmp/f.csv}\\
                               & {\scriptsize\bfseries OBS} & \textit{``/bin/sh: ssconvert: command not found''}
\end{tabular}
\end{casephase}

{\scriptsize\color{CaseFail}
$\blacktriangleright$
No step in this trace has \texttt{action\_type=ROLLBACK}: the agent never requests recovery.
The re-execution at idx 4 of the command already issued at idx 1, returning byte-identical
output, is what the reviewer registers as the attempt having stopped learning.
}

\begin{casephase}{CaseRollback}{CaseRollbackBG}{Horizon-triggered recovery}{$C_6 \to C_2$}
{\scriptsize\bfseries REVIEWER}\;
\textit{``The attempt has stopped learning; repeated actions with no new observations or progress.
The environment lacks the necessary tools (pip, ssconvert) to proceed \ldots''}
\hfill{\scriptsize\color{black!55}(\texttt{HORIZON\_REACHED}, urgency 0.8)}\\[3pt]

{\scriptsize\bfseries SELECTOR}\;
\textit{``The earliest checkpoint (index 2) where the issue was first encountered (pip not found)
is the latest point that still escapes the failure.''}
\hfill{\scriptsize\color{black!55}(range $[2,2]$, single candidate)}
\end{casephase}

\begin{casephase}{CaseMemory}{CaseMemoryBG}{Reflection Memory retained across rollback}{1 entry}
{\scriptsize\bfseries AFFORDANCE}\;
Bash command execution: Requires the command to be available in the environment
(e.g., pip, ssconvert).\\[2pt]

{\scriptsize\bfseries INVALIDATED}\;
\textit{``The spreadsheet can be directly read or converted using available tools
in the current environment.''}
\end{casephase}

\begin{casephase}{CaseRecover}{CaseRecoverBG}{Recovered branch}{idx 2--4}
\begin{tabular}{@{}r@{\hspace{0.6em}}l@{\hspace{0.5em}}p{\dimexpr\linewidth-4.2em\relax}@{}}
{\scriptsize\color{black!55}2} & {\scriptsize\bfseries ACT} & \texttt{unzip -p "\$ZIP" "CATEGORIES.xml"}\\
                               & {\scriptsize\bfseries OBS} & \texttt{<w:wordDocument xmlns:aml=\ldots\ xmlns:wpc=\ldots}
                               {\rm(WordML boilerplate; the}
                               \textsc{categories} {\rm text elements are not reached)}\\[3pt]
{\scriptsize\color{black!55}3} & {\scriptsize\bfseries ACT} & \texttt{unzip -p "\$ZIP" "food\_duplicates.xls" \textgreater{} /tmp/f.xls \&\& strings /tmp/f.xls \textbar{} grep -v "\textbackslash{}x00"}\\
                               & {\scriptsize\bfseries OBS} & \ldots\ chickpea garbanzo bean\ aubergine eggplant\
                               zucchini courgette \ldots\ congee rice porridge\
                               \textbf{turtle soup}\ granola\ tofu soy curds \ldots\\[3pt]
{\scriptsize\color{black!55}4} & {\scriptsize\bfseries ACT} & {[}FINAL{]} \textit{``\ldots\ ``Soups and Stews'', which includes ``turtle soup''.''}\\
                               & {\scriptsize\bfseries OBS} & Final answer produced.
\end{tabular}
\end{casephase}

{\scriptsize\color{CaseRecover}
$\blacktriangleright$
The recovered branch does not repair the parsing toolchain---it abandons it.
\texttt{strings} extracts the paired food names directly from the binary spreadsheet,
leaving \texttt{turtle soup} as the unpaired entry.
}
\hfill
\textcolor{CaseSuccess}{\textbf{\checkmark\ success}}

\begin{casephase}{CaseNeutral}{CaseNeutralBG}{Note on necessity}{}
Unlike G1/G2, the failed commands here did not corrupt any environment state---they simply failed.
An agent that switched strategy \emph{without} rolling back could plausibly have succeeded too.
Rollback's benefit in this case is discarding a context saturated with four dead-end attempts,
not undoing a side effect.
\end{casephase}

\end{casestudy}

\captionof{figure}{
\textbf{G3.}
A GAIA recovery triggered by the adaptive review horizon rather than by an agent-issued
rollback request. After several unavailable parsing tools lead to the same technical dead end,
recovery returns execution to an earlier checkpoint and the next branch adopts a different,
dependency-free strategy.
}
\label{fig:case_g3}

\textbf{Analysis.}
G3 illustrates the complementary role of scheduled review when failure is not recognized
by the acting agent itself. No \texttt{ROLLBACK} action is issued during the failed branch:
the agent continues trying increasingly redundant parsing routes, including re-running a
command that produces an identical observation. The horizon-triggered reviewer detects this
loss of information gain and abandons the tool-dependent parsing strategy. After recovery,
the agent switches to dependency-free shell operations and extracts the required evidence
without repairing the unavailable parsing toolchain. Unlike G1 and G2, rollback does not
undo a persistent environmental side effect in this case; its primary benefit is to terminate
an unproductive branch and reconstruct the decision context around a different strategy.
G3 therefore mainly demonstrates the value of RIR's external review channel rather than the
strict necessity of physical state restoration.


\subsection{Bad Cases}
\label{sec:case_bad}


\subsubsection{B1 --- ALFWorld, incorrect reflection persists across attempts}

\begin{casestudy}{B1 \quad ALFWorld \quad \textcolor{CaseCrit}{\textbf{Failure}}}

{\small\textbf{Task.}
\texttt{put a clean cloth in countertop.}
}


\begin{casephase}
{CaseNeutral}
{CaseNeutralBG}
{Attempt 1 --- the failure condition is created}
{idx 0--17}

\begin{tabular}{@{}r@{\hspace{0.6em}}l@{\hspace{0.5em}}p{\dimexpr\linewidth-4.2em\relax}@{}}
{\scriptsize\color{black!55}0} & {\scriptsize\bfseries ACT} & \texttt{go to handtowelholder 1}\\
                               & {\scriptsize\bfseries OBS} & You arrive at handtowelholder 1. On the handtowelholder 1, you see a handtowel 1.\\[3pt]
{\scriptsize\color{black!55}1} & {\scriptsize\bfseries ACT} & \texttt{take handtowel 1 from handtowelholder 1}\\
                               & {\scriptsize\bfseries OBS} & You pick up the handtowel 1 from the handtowelholder 1.
                               \hfill{\scriptsize\color{CaseFail}(the single hand slot is now occupied)}\\[3pt]
{\scriptsize\color{black!55}2} & {\scriptsize\bfseries ACT} & \texttt{go to sinkbasin 1}\\
                               & {\scriptsize\bfseries OBS} & You arrive at sinkbasin 1. On the sinkbasin 1, you see a soapbar 1.\\[3pt]
{\scriptsize\color{black!55}3} & {\scriptsize\bfseries ACT} & \texttt{clean handtowel 1 with sinkbasin 1}\\
                               & {\scriptsize\bfseries OBS} & \textit{``Nothing happens.''}\\[3pt]
{\scriptsize\color{black!55}4} & {\scriptsize\bfseries ACT} & {[}ROLLBACK requested{]} \casebadge{CaseBad}{denied}\\[3pt]
{\scriptsize\color{black!55}6} & {\scriptsize\bfseries ACT} & \texttt{take soapbar 1 from sinkbasin 1}
                               \hfill{\scriptsize\color{black!45}(hand still holds the handtowel)}\\
                               & {\scriptsize\bfseries OBS} & \textit{``Nothing happens.''}\\[3pt]
{\scriptsize\color{black!55}10} & {\scriptsize\bfseries ACT} & \texttt{go to countertop 1}\\
                                & {\scriptsize\bfseries OBS} & On the countertop 1, you see a candle 1,
                                \textbf{a cloth 3}, and a spraybottle 1.
                                \hfill{\scriptsize\color{CaseRecover}(the task target is present)}\\[3pt]
{\scriptsize\color{black!55}11} & {\scriptsize\bfseries ACT} & \texttt{take cloth 3 from countertop 1}\\
                                & {\scriptsize\bfseries OBS} & \textit{``Nothing happens.''}\\[3pt]
\multicolumn{3}{@{}l@{}}{\scriptsize\color{black!45}\textit{idx 5, 7--9, 12--16 omitted:
further \texttt{take}/\texttt{examine} attempts on towel 1, spraybottle 1, cloth 3 ---
all \textit{``Nothing happens.''}}}\\[3pt]
{\scriptsize\color{black!55}17} & {\scriptsize\bfseries ACT} & {[}ROLLBACK requested{]}
                                \hfill{\scriptsize\color{black!45}(approved: rollback \#1)}\\
                                & {\scriptsize\bfseries WHY} & \textit{``After multiple attempts, I have not found a way to clean t\ldots''}
\end{tabular}
\end{casephase}

{\scriptsize\color{CaseFail}
$\blacktriangleright$
The critical state change occurs at idx 1: the hand becomes occupied before
any subsequent interaction is attempted.
Every later \texttt{take} in this attempt returns the same undifferentiated
signal \textit{``Nothing happens.''}, which the environment also emits for
inapplicable actions---the agent cannot distinguish a full hand
from a non-interactable object.
}

\begin{casephase}
{CaseBad}
{CaseBadBG}
{First diagnosis}
{lesson emitted at idx 17}

\textit{
``Cleaning the handtowel with the sinkbasin failed because the sinkbasin
lacked the required cleaning capability (no observable cleaning effect).
Taking the soapbar from the sinkbasin failed, possibly due to unmet
preconditions (e.g., the soapbar was not interactable).''
}

\end{casephase}

{\scriptsize\color{CaseBad}
$\blacktriangleright$
The review reaches the correct abstraction---\emph{unmet preconditions}---
but instantiates it on the sinkbasin and the target objects rather than on
the agent's occupied hand.
The same run contains 80 successful \texttt{clean \ldots\ with sinkbasin}
actions across all controllers, so the sinkbasin does possess the capability
the review denies it.
}


\begin{casephase}
{CaseRollback}
{CaseRollbackBG}
{Early recovery cycles}
{rollback \#1: $C_{17}\to C_2$ \quad rollback \#2: $C_{13}\to C_2$}

{\scriptsize\bfseries \#1}\;
\textit{``Restoring to step 2 (going to sinkbasin 1) avoids the failed cleaning action
and allows reassessment of the cleaning approach without repeating the failure.''}\\[3pt]

{\scriptsize\bfseries \#2}\;
\textit{``Checkpoint 2 is the latest checkpoint that avoids the failed cleaning
action with the sinkbasin (step 3) and \textbf{retains the handtowel}
for alternative cleaning attempts.''}

\end{casephase}

{\scriptsize\color{CaseRollback}
$\blacktriangleright$
Both rollbacks remove the failed cleaning action, but restore a checkpoint
at which the agent is still holding the handtowel.
Rollback \#2 makes this explicit: retaining the handtowel is treated as
preserving sound work. The blocking condition therefore survives the early recoveries.
}


\begin{casephase}
{CaseRollback}
{CaseRollbackBG}
{Attempt 4 --- rollback finally precedes the true cause}
{rollback \#4: $C_{13}\to C_0$, range $[0,1]$}

\textit{
``Restoring to checkpoint 0 \textbf{(before taking the handtowel)} allows exploring
alternative methods to clean the cloth, as checkpoint 1 leads directly to the
failure due to the lack of viable cleaning tools.''
}

\end{casephase}

{\scriptsize\color{CaseRollback}
$\blacktriangleright$
This selector rationale correctly identifies acquisition of the handtowel
as the action that must be undone and restores a state in which the hand is empty.
It also contrasts with rollback \#2, which preserved the same action as sound work.
}

\begin{casephase}
{CaseFail}
{CaseFailBG}
{Attempt 4 --- the failure nevertheless reappears}
{after restoring $C_0$}

\begin{tabular}{@{}r@{\hspace{0.6em}}l@{\hspace{0.5em}}p{\dimexpr\linewidth-4.2em\relax}@{}}
{\scriptsize\color{black!55}0} & {\scriptsize\bfseries ACT} & \texttt{go to handtowelholder 1}\\
                               & {\scriptsize\bfseries OBS} & On the handtowelholder 1, you see a handtowel 1.\\[3pt]
{\scriptsize\color{black!55}1} & {\scriptsize\bfseries ACT} & \texttt{take handtowel 1 from handtowelholder 1}
                               \hfill{\scriptsize\color{CaseCrit}(the blocking action, re-issued verbatim)}\\
                               & {\scriptsize\bfseries OBS} & You pick up the handtowel 1 from the handtowelholder 1.\\[3pt]
{\scriptsize\color{black!55}2} & {\scriptsize\bfseries ACT} & \texttt{go to sinkbasin 1}\\
                               & {\scriptsize\bfseries OBS} & On the sinkbasin 1, you see a soapbar 1.\\[3pt]
{\scriptsize\color{black!55}3} & {\scriptsize\bfseries ACT} & \texttt{clean handtowel 1 with sinkbasin 1}\\
                               & {\scriptsize\bfseries OBS} & \textit{``Nothing happens.''}\\[3pt]
{\scriptsize\color{black!55}4} & {\scriptsize\bfseries ACT} & {[}ROLLBACK requested{]}
                               \hfill{\scriptsize\color{black!45}(approved: rollback \#5, $C_4\to C_2$)}
\end{tabular}
\end{casephase}

{\scriptsize\color{CaseCrit}
$\blacktriangleright$
The physical rollback is now deep enough: the agent begins again with an empty hand and
the original dead end is no longer entailed by the restored environment state.
It nevertheless re-issues the blocking action immediately, while the persistent memory
continues to discourage \texttt{cloth 3}.
}


\begin{casephase}
{CaseCrit}
{CaseCritBG}
{Collapse into a two-step limit cycle}
{rollbacks \#5 and \#6, both $C_4\to C_2$, depth 2}

\begin{tabular}{@{}r@{\hspace{0.6em}}l@{\hspace{0.5em}}p{\dimexpr\linewidth-4.2em\relax}@{}}
{\scriptsize\color{black!55}2} & {\scriptsize\bfseries ACT} & \texttt{go to sinkbasin 1}\\
{\scriptsize\color{black!55}3} & {\scriptsize\bfseries ACT} & \texttt{clean handtowel 1 with sinkbasin 1}
                               $\;\rightarrow\;$ \textit{``Nothing happens.''}\\
{\scriptsize\color{black!55}4} & {\scriptsize\bfseries WHY} & \textit{``Cleaning the handtowel with the sinkbasin failed,
                               \textbf{as established in earlier attempts}\ldots''}\\[3pt]
\multicolumn{3}{@{}l@{}}{\scriptsize\color{CaseCrit}$\hookrightarrow$
restored to $C_2$; the identical two steps are issued again}
\end{tabular}
\end{casephase}

{\scriptsize\color{CaseCrit}
$\blacktriangleright$
The agent states that the action is already known to fail and issues it regardless.
Restored to $C_2$, it stands at the sinkbasin holding the handtowel, while its retained
knowledge provides no productive alternative.
}


\begin{casephase}
{CaseMemory}
{CaseMemoryBG}
{Reflection Memory accumulated across failed attempts}
{5 of 7 entries}

\begin{tabular}{@{}r@{\hspace{0.6em}}p{\dimexpr\linewidth-2.6em\relax}@{}}
{\scriptsize\color{black!55}3} & taking an item: requires the item to be interactable
(e.g., \textbf{the soapbar 1 and cloth 3} are not interactable)\\[2pt]
{\scriptsize\color{black!55}4} & taking an item: requires the item to be interactable
(e.g., \textbf{the spraybottle 1 and cloth 3} are not interactable)\\[2pt]
{\scriptsize\color{black!55}5} & taking an item: requires the item to be interactable
(e.g., \textbf{the cloth 3 and spraybottle 1} are not interactable)
\hfill{\scriptsize\color{CaseCrit}(same content as \#4, order reversed)}\\[2pt]
{\scriptsize\color{black!55}6} & taking an item: requires the item to be interactable
(e.g., \textbf{the cloth 3} is not interactable)\\[2pt]
{\scriptsize\color{black!55}7} & taking an item: requires the item to be interactable
(e.g., \textbf{the towel 1} is not interactable)
\hfill{\scriptsize\color{CaseCrit}(logged with \texttt{restated=1})}
\end{tabular}
\end{casephase}

{\scriptsize\color{CaseMemory}
$\blacktriangleright$
\texttt{cloth 3}---the correct task target---is recorded as non-interactable
in four of these five entries.
Entries \#4 and \#5 carry identical information and differ only in the order in
which the two objects are listed; the deduplication key is literal, so both are
admitted. Entry \#7 is flagged as a restatement and retained anyway.
Persistent memory therefore increasingly reinforces the incorrect interpretation
formed during the failed attempts.
}

\begin{casephase}
{CaseCrit}
{CaseCritBG}
{Restore-point repetition across the episode}
{6 of 9 rollbacks return to $C_2$}

\begin{tabular}{@{}c@{\hspace{0.9em}}c@{\hspace{0.9em}}c@{\hspace{0.9em}}p{\dimexpr\linewidth-9.5em\relax}@{}}
{\scriptsize\bfseries \#} & {\scriptsize\bfseries range} & {\scriptsize\bfseries $\to$} &
{\scriptsize\bfseries selector's stated reason}\\[2pt]
{\scriptsize 2} & {\scriptsize $[1,2]$} & {\scriptsize 2} &
\textit{``lower \texttt{prior\_rollbacks} count \textbf{(1)} compared to checkpoint 1 \textbf{(0)}''}\\[2pt]
{\scriptsize 3} & {\scriptsize $[2,4]$} & {\scriptsize 2} &
\textit{``lowest \texttt{prior\_rollbacks} count \textbf{(2)} compared to other candidates''}\\[2pt]
{\scriptsize 7} & {\scriptsize $[1,2]$} & {\scriptsize 2} &
\textit{``lower \texttt{prior\_rollbacks} count compared to checkpoint 1''}
\end{tabular}
\end{casephase}

{\scriptsize\color{CaseCrit}
$\blacktriangleright$
The field is intended to discourage repeatedly restoring the same checkpoint.
In these examples, however, the selector treats the larger prior-rollback count as preferable,
contributing to repeated restoration of $C_2$.
}


\begin{casephase}
{CaseCrit}
{CaseCritBG}
{Outcome}
{}

Budget exhausted at 100 steps.
Across the episode: 8 executions of \texttt{clean handtowel 1 with sinkbasin 1},
5 of \texttt{take cloth 3 from countertop 1}, all returning
\textit{``Nothing happens.''};
checkpoint revisit rate 44.44\%, no-op patch rate 36.36\%,
restated-knowledge rate 41.67\%.
\textcolor{CaseCrit}{
\textbf{\ding{55}\ failure}
}

\end{casephase}

\end{casestudy}

\captionof{figure}{
\textbf{B1.}
An early diagnostic error propagates across successive recovery attempts.
Initial rollbacks preserve the occupied-hand state responsible for the failure, while repeated
failed interactions introduce misleading claims about the valid target into Reflection Memory.
Even when rollback \#4 finally restores a state before the original blocking action, the
retained knowledge helps reconstruct the same failed branch, after which execution collapses
into a short recovery cycle.
}
\label{fig:case_b1}

\textbf{Analysis.}
B1 exposes a compounding failure across both restore localization and persistent reflection.
The initial observation, \textit{``Nothing happens,''} is ambiguous, and the reviewer
incorrectly attributes the failure to the sinkbasin or target objects rather than to the
agent's occupied hand. Early rollbacks therefore preserve the handtowel as valid progress,
while repeated failed interactions are progressively generalized into Reflection Memory
entries that mark \texttt{cloth 3} as non-interactable. Restore localization also becomes
increasingly repetitive, with six of nine rollbacks returning to $C_2$.

Rollback \#4 isolates the role of memory particularly clearly. It correctly restores $C_0$,
before the handtowel is acquired, and therefore removes the original blocking state.
Yet the agent immediately reconstructs the same branch while the retained Reflection Memory
discourages the valid cloth target. B1 therefore shows that successful rollback requires
more than reaching a causally valid checkpoint: the knowledge carried across the rollback
boundary must also remain consistent with the alternatives available in the restored state.


\subsubsection{B2 --- ScienceWorld, recovery without a concrete repair}

\begin{casestudy}{B2 \quad ScienceWorld \quad \textcolor{CaseCrit}{\textbf{Failure}}}

{\small\textbf{Task.}
\texttt{
Use chemistry to create `red paper'. A recipe and some ingredients might be found near the workshop.
When done, focus on the red paper.
}
}


\begin{casephase}
{CaseNeutral}
{CaseNeutralBG}
{Attempt 1 --- the recipe is obtained and read}
{idx 0--5}

\begin{tabular}{@{}r@{\hspace{0.6em}}l@{\hspace{0.5em}}p{\dimexpr\linewidth-4.2em\relax}@{}}
{\scriptsize\color{black!55}0} & {\scriptsize\bfseries ACT} & \texttt{teleport to workshop}\\
                               & {\scriptsize\bfseries OBS} & You teleport to the workshop.\\[3pt]
{\scriptsize\color{black!55}1} & {\scriptsize\bfseries ACT} & \texttt{look around}\\
                               & {\scriptsize\bfseries OBS} & \ldots\ you see: the agent\ a substance called air\
                               \textbf{A blank paper}\ \textbf{A recipe titled instructions to make red paper}\
                               a table. On the table is: a battery, \ldots\ a ultra low temperature freezer.\\[3pt]
{\scriptsize\color{black!55}2} & {\scriptsize\bfseries ACT} & \texttt{pick up recipe}
                               \hfill{\scriptsize\color{CaseRecover}(short form, not the displayed name)}\\
                               & {\scriptsize\bfseries OBS} & \textit{``You move the recipe to the inventory.''}
                               \hfill{\scriptsize\color{CaseRecover}(succeeds)}\\[3pt]
{\scriptsize\color{black!55}3} & {\scriptsize\bfseries ACT} & \texttt{read recipe in inventory}\\
                               & {\scriptsize\bfseries OBS} & \textit{``The recipe reads: To make red paper,
                               you need to \textbf{mix} paper, red paint.''}\\[3pt]
{\scriptsize\color{black!55}4} & {\scriptsize\bfseries ACT} & \texttt{look around}
                               \hfill{\scriptsize\color{black!45}(repeated)}\\[3pt]
{\scriptsize\color{black!55}5} & {\scriptsize\bfseries ACT} & \texttt{pick up blank paper}\\
                               & {\scriptsize\bfseries OBS} & \textit{``No known action matches that input.''}
\end{tabular}
\end{casephase}

{\scriptsize\color{CaseNeutral}
$\blacktriangleright$
Two facts are established in the first four steps and never revised.
The recipe explicitly names the required operation, \texttt{mix}.
Moreover, \texttt{pick up recipe} succeeds at idx 2 using a short form rather than the
displayed name \textit{``A recipe titled instructions to make red paper''},
showing that short-form reference can work in this environment.
}


\begin{casephase}
{CaseBad}
{CaseBadBG}
{Early reflection focuses on object reference}
{across recovery cycles}

\textit{
``Interacting with `A blank paper' failed due to unrecognized inputs
while in the workshop.''
}\\[3pt]

Reflection Memory:
\textit{
``picking up an item: requires the item to be present and accessible
\textbf{with the correct name}'';
``interacting with paper: requires the correct action name or method to interact.''
}

\end{casephase}

{\scriptsize\color{CaseBad}
$\blacktriangleright$
The diagnosis remains at the level of an invalid name or method.
It does not instantiate the concrete alternative already present in the recipe:
\texttt{mix}.
}


\begin{casephase}
{CaseNeutral}
{CaseNeutralBG}
{Attempt 8 --- the missing ingredient is acquired}
{idx 15--23}

\begin{tabular}{@{}r@{\hspace{0.6em}}l@{\hspace{0.5em}}p{\dimexpr\linewidth-4.2em\relax}@{}}
{\scriptsize\color{black!55}15} & {\scriptsize\bfseries ACT} & \texttt{teleport to art studio}\\
                                & {\scriptsize\bfseries OBS} & You teleport to the art studio.\\[3pt]
{\scriptsize\color{black!55}17} & {\scriptsize\bfseries ACT} & \texttt{pick up wood cup containing red paint}\\
                                & {\scriptsize\bfseries OBS} & \textit{``You move the wood cup to the inventory.''}
                                \hfill{\scriptsize\color{CaseRecover}(succeeds)}\\[3pt]
{\scriptsize\color{black!55}21} & {\scriptsize\bfseries ACT} & \texttt{move wood cup containing red paint to table}\\
                                & {\scriptsize\bfseries OBS} & \textit{``You move the wood cup to the table.''}\\[3pt]
{\scriptsize\color{black!55}23} & {\scriptsize\bfseries ACT} & \texttt{move A blank paper to wood cup containing red paint}\\
                                & {\scriptsize\bfseries OBS} & \textit{``No known action matches that input.''}
\end{tabular}
\end{casephase}

{\scriptsize\color{CaseNeutral}
$\blacktriangleright$
The red paint is located in the art studio rather than the workshop.
The agent successfully acquires it and returns it to the relevant setting.
At this point the recipe and required ingredient are available, but the synthesis
operation itself remains untried.
}


\begin{casephase}
{CaseRollback}
{CaseRollbackBG}
{Recovery cycles repeatedly return to the recipe}
{7 of 14 rollbacks restore idx 3}

\begin{tabular}{@{}l@{\hspace{1em}}l@{}}
{\scriptsize\bfseries restore target}
& {\scriptsize idx 3 ($\times 7$)\quad idx 2 ($\times 2$)\quad idx 4 ($\times 2$)\quad idx 0, 17, 18 ($\times 1$ each)}\\[2pt]
{\scriptsize\bfseries recurrent pattern}
& {\scriptsize rollback $\rightarrow$ restore near the recipe
$\rightarrow$ retry a paper reference $\rightarrow$ unrecognized}\\
& {\scriptsize $\rightarrow$ reflect on ``correct name or method''
$\rightarrow$ rollback}
\end{tabular}
\end{casephase}

{\scriptsize\color{CaseRollback}
$\blacktriangleright$
Restoring to idx 3 is itself a defensible choice: it retains the recipe while
discarding failed interactions. However, the information carried into the next
attempt continues to prioritize the reference hypothesis rather than a new operation.
Of the 150 executed steps, 68 are replays of previously executed actions.
}


\begin{casephase}
{CaseFail}
{CaseFailBG}
{Reference search dominates the episode}
{full-trajectory aggregate}

\begin{tabular}{@{}r@{\hspace{0.9em}}l@{\hspace{0.9em}}p{\dimexpr\linewidth-11em\relax}@{}}
{\scriptsize\bfseries count} & {\scriptsize\bfseries action} & {\scriptsize\bfseries result}\\[2pt]
{\scriptsize 14} & \texttt{look at blank paper}      & \textit{``No known action matches that input.''}\\
{\scriptsize 10} & \texttt{look at A blank paper}    & \textit{same}\\
{\scriptsize 12} & \texttt{pick up blank paper}      & \textit{same}\\
{\scriptsize\phantom{0}3} & \texttt{pick up A blank paper} & \textit{same}\\[2pt]
\multicolumn{3}{@{}l@{}}{\scriptsize\color{CaseFail}$=$ 39 attempts spanning both verbs $\times$ both spellings of the name}\\[4pt]
{\scriptsize\phantom{0}0} & \texttt{mix \ldots}       & \textbf{never issued}
\end{tabular}
\end{casephase}

{\scriptsize\color{CaseFail}
$\blacktriangleright$
The agent repeatedly varies the verb and spelling of the paper reference.
The operation explicitly named in the recipe---which is read thirteen times over
the episode---is never attempted.
}


\begin{casephase}
{CaseMemory}
{CaseMemoryBG}
{Reflection Memory grows without adding an actionable hypothesis}
{9 entries, restated rate 66.67\%}

\begin{tabular}{@{}r@{\hspace{0.6em}}p{\dimexpr\linewidth-2.6em\relax}@{}}
{\scriptsize\color{black!55}1} & picking up an item: requires the item to be present and accessible\\[2pt]
{\scriptsize\color{black!55}2} & picking up an item: requires the item to be present and accessible
\textbf{in a specific state or location}\\[2pt]
{\scriptsize\color{black!55}3} & picking up an item: requires the item to be present and accessible
\textbf{with the correct name}
\hfill{\scriptsize\color{CaseCrit}(each entry extends the previous one)}\\[4pt]
{\scriptsize\color{black!55}8} & workshop: holds exactly: a blank paper, a table with \ldots\
a switch (off), \textbf{a yellow light bulb (off), a wood cup (containing red paint)}\\[2pt]
{\scriptsize\color{black!55}9} & workshop: holds exactly: a blank paper, a table with \ldots\
a switch (off), \textbf{a wood cup (containing red paint), a yellow light bulb (off)}
\hfill{\scriptsize\color{CaseCrit}(identical content, last two items transposed)}
\end{tabular}
\end{casephase}

{\scriptsize\color{CaseMemory}
$\blacktriangleright$
Entries 1--3 restate the same general hypothesis with increasing specificity,
while entries 8 and 9 differ only in list order.
The recorded restatement rate is 66.67\%: the memory grows, but the set of
actionable alternatives does not.
}


\begin{casephase}
{CaseCrit}
{CaseCritBG}
{Outcome}
{}

Budget exhausted at 150 steps
(\texttt{termination\_reason = max\_total\_steps}).
The recipe is read 13 times.
The agent issues \texttt{look at}, \texttt{pick up}, \texttt{pour},
\texttt{use \ldots\ on}, and \texttt{move \ldots\ to} against the paper in
multiple formulations---39 reference attempts in total---
and \textbf{never issues \texttt{mix}}.
No-op patch rate 64.29\%, checkpoint revisit rate 35.71\%.
\textcolor{CaseCrit}{
\textbf{\ding{55}\ failure, reward 0.33}
}
\textit{
(partial credit consistent with the recipe located and the ingredient acquired,
synthesis not completed)
}

\end{casephase}

\end{casestudy}

\captionof{figure}{
\textbf{B2.}
The agent obtains a recipe that explicitly specifies \texttt{mix} and later acquires
the required red paint, yet subsequent recovery cycles continue to focus on how the paper
should be referenced. Across the full episode, the agent tries 39 reference variants but
never issues the operation provided by the recipe.
}
\label{fig:case_b2}

\textbf{Analysis.}
B2 illustrates a different failure mode from B1: the system repeatedly diagnoses the problem
at the wrong level of abstraction. After the first failed paper interaction, reflection focuses
on finding the ``correct name or method.'' Yet the trajectory already contains evidence
against a pure naming explanation---\texttt{pick up recipe} succeeds at idx 2 despite using
a shortened reference---and the recipe itself explicitly supplies the missing operation,
\texttt{mix}. Later attempts also acquire the red paint, leaving the synthesis operation as
the unresolved step. Nevertheless, recovery repeatedly returns to the same reference-search
hypothesis, producing 39 naming variants and no \texttt{mix} action.

Unlike B1, the restored checkpoints themselves are generally reasonable. The failure lies
primarily in the information carried forward: Reflection Memory repeatedly elaborates the
same hypothesis without producing a concrete alternative that changes the next action.
B2 therefore shows that useful reflection must be not only consistent with the trajectory,
but actionable enough to alter the decision process after recovery.


\subsection{Summary of the Case Studies}
\label{sec:case_summary}

Across the successful trajectories, the three components of RIR play complementary roles.
G1 demonstrates why explicit state restoration is valuable once an erroneous action has
already changed the environment, while Reflection Memory prevents useful exploration from
being discarded with that state. G2 shows that restoring an earlier checkpoint can recover
an execution precondition that textual feedback alone does not repair. G3 complements these
examples on the \emph{when} dimension: scheduled review can terminate an unproductive branch
even when the acting agent never requests recovery. Together, these cases illustrate the
advantage of treating recovery as a joint decision over \emph{when} to intervene,
\emph{where} to resume, and \emph{what} information should survive.

The failed trajectories expose the corresponding requirements. B1 shows that incorrect
failure attribution can propagate through both restore selection and persistent memory,
remaining harmful even after execution is eventually restored to an appropriate state.
B2 shows that a broadly plausible reflection may still be ineffective when it fails to
produce an actionable alternative. Effective recovery therefore requires state restoration
and information retention to reinforce one another: rollback must remove the condition that
made the previous branch fail, while reflection must make the next attempt meaningfully
different. This is the central advantage targeted by RIR over either mechanism in isolation:
execution can return to a useful prior state without discarding reusable knowledge acquired
along the abandoned branch.

\end{document}